\documentclass[journal]{IEEEtran}
\usepackage[pdftex]{graphicx}
\graphicspath{{images/}}
\usepackage{algorithmic}
\usepackage{array}
\usepackage [autostyle, english = american]{csquotes}
\MakeOuterQuote{"}
\usepackage{soul}
\usepackage[caption=false,font=normalsize,labelfont=sf,textfont=sf]{subfig}
\newcommand\figref{Fig.~\ref}
\newcommand\etal{\textit{et al. }}
\usepackage{makecell}
\bstctlcite{IEEEexample:BSTcontrol}
\usepackage{url}
\usepackage{color}
\usepackage{subfig}
\usepackage{hyperref}
\hypersetup{
    colorlinks=true,
    linkcolor=blue,
    filecolor=magenta,      
    urlcolor=cyan,
    citecolor=blue,
    breaklinks=true
}

\usepackage{csquotes}
\usepackage[utf8]{inputenc}
\usepackage{marginnote}

\begin{document}
\title{Autonomous Vehicles that Interact with Pedestrians: A Survey of Theory and Practice}
\author{Amir~Rasouli and John K.~Tsotsos
\thanks{The authors are with the Department of Electrical Engineering and Computer Science, York University, Toronto, ON, Canada, e-mail: \tt\small \{aras,tsotsos\}@eecs.yorku.ca}\\
}

\maketitle
\begin{abstract}
One of the major challenges that autonomous cars are facing today is driving in urban environments. To make it a reality, autonomous vehicles require the ability to communicate with other road users and understand their intentions. Such interactions are essential between the vehicles and pedestrians as the most vulnerable road users. Understanding pedestrian behavior, however, is not intuitive and depends on various factors such as demographics of the pedestrians, traffic dynamics, environmental conditions, etc. In this paper, we identify these factors by surveying  pedestrian behavior studies, both the classical works on pedestrian-driver interaction and the modern ones that involve autonomous vehicles. To this end, we will discuss various methods of studying pedestrian behavior, and analyze how the factors identified in the literature are interrelated. We will also review the practical applications aimed at solving the interaction problem including design approaches for autonomous vehicles that communicate with pedestrians and visual perception and reasoning  algorithms tailored to understanding pedestrian intention.  Based on our findings, we will discuss the  open problems and propose future research directions.
\end{abstract}
\begin{IEEEkeywords}
Autonomous vehicles, Pedestrian behavior, Traffic interaction, Survey.
\end{IEEEkeywords}
\vspace*{-0.2cm} 
\section{Introduction}

Ever since the introduction of early commercial automobiles, engineers and scientists have been striving to achieve autonomy, that is removing the need for human involvement in controlling the vehicles. Apart from the increased level of comfort for drivers, autonomous vehicles can positively impact society both at micro and macro levels \cite{winkle2016safety,litman2014autonomous}. 

Replacing human drivers with autonomous control systems, however, comes at he price of creating a social interaction void. Besides being a dynamic control task, driving is a social phenomenon and requires interactions between all road users involved to ensure the flow of traffic and to guarantee the safety of others \cite{rasouli2017understanding}. 

\begin{figure}[!tp]
\centering
\includegraphics[width=0.9\columnwidth]{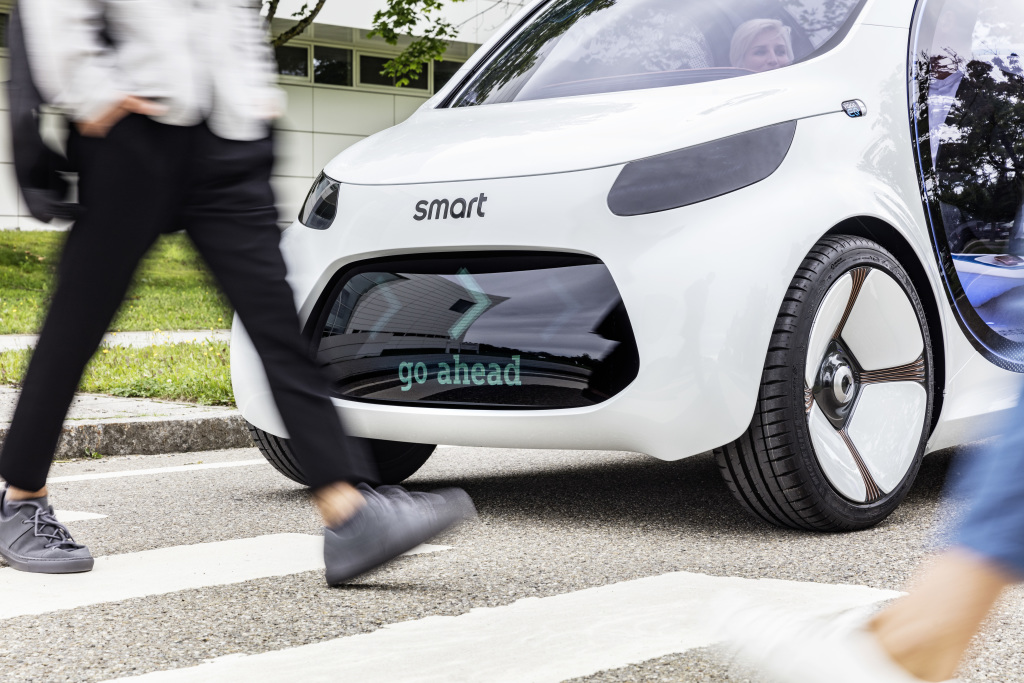}
\caption[smart]{The autonomous car is communicating with pedestrians at a crosswalk indicating that it is safe to cross. Source: \cite{smartcomm}.}
\vspace*{-0.5cm} 

\label{fig:smart}
\end{figure}

Social interaction can play an important role in resolving various potential ambiguities in traffic. For example, if a car wants to turn at a non-signalized intersection on a heavily travelled street, it might wait for another driver's signal indicating the right of way. In the case of pedestrians, interaction can help them to understand when it is safe for them to cross the road, e.g. by receiving a signal from the driver \cite{wolf2016interaction} (see \figref{fig:smart}). Recent field studies of autonomous vehicles show how the lack of social understanding can result in traffic accidents \cite{anthony2016} or erratic behaviors towards pedestrians \cite{richtel2016}.

Given that autonomous vehicles may commute without any passengers on board, they are subject to malicious behavior, similar to those observed against a number of autonomous robots used in malls \cite{bully}. For example, some people might step in front of the autonomous vehicles to force them to change their route or interrupt their operation. Understanding the true intention of these people can help the vehicles act accordingly.

A large body of studies in the field of behavioral psychology have addressed the social aspects of driving and identified numerous factors that can potentially influence the way road users behave \cite{ishaque2008behavioural,willis2004human,harrell1991factors}. Factors such as pedestrians' demographics \cite{cohen1955risk}, road conditions \cite{harrell1991factors}, social factors \cite{willis2004human}, and traffic characteristics \cite{jacobs1967study} are shown to significantly influence pedestrian crossing decisions. However, there is a missing component in the literature, namely a holistic view of pedestrian crossing behavior to identify the extent of these factors and to explain in what ways they are interrelated.  

In the context of intelligent driving, intention estimation algorithms have been developed to predict forthcoming actions of pedestrians \cite{rasouli2017they} and drivers \cite{molchanov2015multi}. Technologies have also been introduced that enable autonomous vehicles to communicate with road users, such as V2V \cite{hobert2015enhancements} and V2P \cite{hussein2016p2v} wireless communication mechanisms, and various visual intent displays such as LED lights \cite{lagstrom2015avip} or projectors \cite{benzauto}. The majority of these approaches, nonetheless, disregard the theoretical findings of traffic interaction and treat the problem as dealing with a rigid dynamic object rather than a social being \cite{schulz2015controlled}. 

This paper addresses the above shortcomings and establishes a connection between studies on traffic interaction from different disciplines. More specifically, we first discuss various methods of studying pedestrian behavior, their efficiency and popularity in the literature. We then conduct a comprehensive review of pedestrian behavior studies including the classical studies on driver-pedestrian interactions and the studies that involve autonomous vehicles. Based on our findings we present a visualization highlighting past studies of pedestrian behavior and how they are connected to one another. In the second part of the paper, we focus our attention on the practical systems designed for communicating with pedestrians, and understanding and predicting their behavior. We conclude our paper with discussion of open research problems in the field of traffic social interaction and proposal for future directions.

\section{Methods of study}

The methods of studying human behavior (in traffic scenes) have transformed during past decades as new technological advancements have emerged. Traditionally, written questionnaires \cite{wilde1980immediate,price2000relationship} or direct interviews \cite{crundall1999driving} were widely used to collect information from traffic participants or authorities monitoring the traffic. Some modern studies still rely on questionnaires especially in cases where there is a need to measure the general attitudes of people towards various aspects of driving, e.g. crossing in front of autonomous vehicles \cite{deb2017development}. These forms of studies, however, have been criticized for the bias people have in answering questions, the honesty of participants in responding or even how well the interviewees are able to recall a particular traffic situation.

Traffic reports are mainly generated by professionals such as police forces after accidents \cite{sullivan2011differences}. The advantage of traffic reports is that they provide good detail regarding the elements involved in a traffic accident, albeit not being able to substantiate the underlying reasons.

In addition, behavior can be analyzed via on-site observation by the researcher either present in the vehicle \cite{risser1985behavior} or standing outside \cite{tom2011gender} while recording the behavior of the road users. Observations can be both naturalistic and scripted. In a naturalistic format, normal activities of road users are monitored without notifying them of such recording \cite{lefkowitz1955status}. In a scripted setting, the participants, e.g. drivers or pedestrians, are instructed to perform certain actions, and then the reactions of other parties are observed \cite{schmidt2009pedestrians,dey2017impact}. A major drawback of observation is the strong observer bias, which can be caused by both the observers' misperception of the traffic scenes or their subjective judgments.

New technological developments in the design of sensors and cameras have given rise to different modalities of recording traffic events. Eye tracking devices are one such system that can record participants' eye movements during driving \cite{clay1995driver}. Computer simulations \cite{reed2008intersection} and video recordings of traffic scenes \cite{price2000relationship} are also widely used to study the behavior of drivers in laboratory environments. These methods, however, have been criticized for not providing realistic driving conditions, therefore the observed behaviors may not necessarily reflect the ones exhibited by road users in a real traffic scenario.

Naturalistic recording of traffic scenes (both videos \cite{kotseruba2016joint} and photos \cite{dipietro1970pedestrian}), is, perhaps, one of the most effective methods for studying traffic behavior. Although the first instances of such studies date back to almost half a century ago \cite{heimstra1969experimental}, they have gained tremendous popularity in recent years. In this method of study, a camera (or a network of cameras) are placed either inside the vehicles  \cite{neale2005overview,eenink2014udrive,rasouliagree} or outside on roadsides \cite{sun2002modeling,wang2010study}. Since the objective is to record the natural behavior of the road users, the cameras are located in inconspicuous places not visible to the observees. In the context of recording driving habits, although the presence of the camera might be known to the driver, it does not alter the driver's behavior in the long run. In fact, studies show that the presence of cameras may only influence the first 10-15 minutes of the driving, hence the beginning of each recording is usually discarded at the time of analysis \cite{risser1985behavior}. An added advantage of recording compared to on-site observation is the possibility of revising the observation and using multiple observers to minimize bias \cite{heimstra1969experimental}. 

Naturalistic recording, similar to on-site observation, may also be affected by observer bias. Moreover, in some cases, it is hard to recognize certain behaviors or underlying motives, e.g. whether a pedestrian notices the presence of the car or looks at the traffic signal in the scene and why. To remedy this issue, it is common to employ a hybrid approach where recordings or observations are combined with on-site interviews \cite{sucha2017pedestrian}. Using this method, after recording a behavior, the researcher approaches the corresponding road user and asks questions regarding their experience, for example, whether they looked at the signal prior to crossing. Overall, the hybrid approach can help resolve the ambiguities observed in certain behaviors.

\begin{figure}[!tp]
\centering
\subfloat[]{
\includegraphics[width=2.5cm,height=3.25cm]{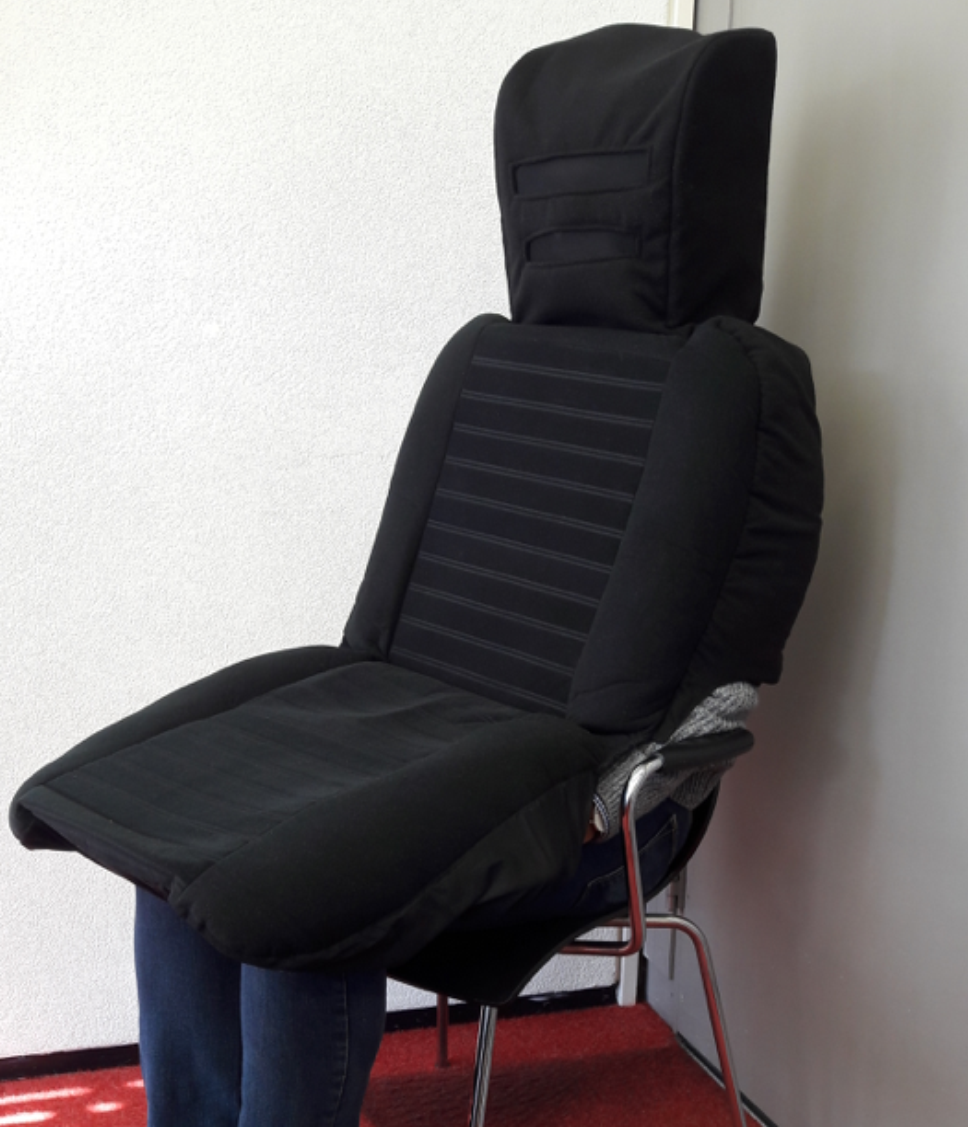}
\label{fig:driver_disguise} }
\subfloat[]{
\includegraphics[width=4.75cm,height=3.25cm]{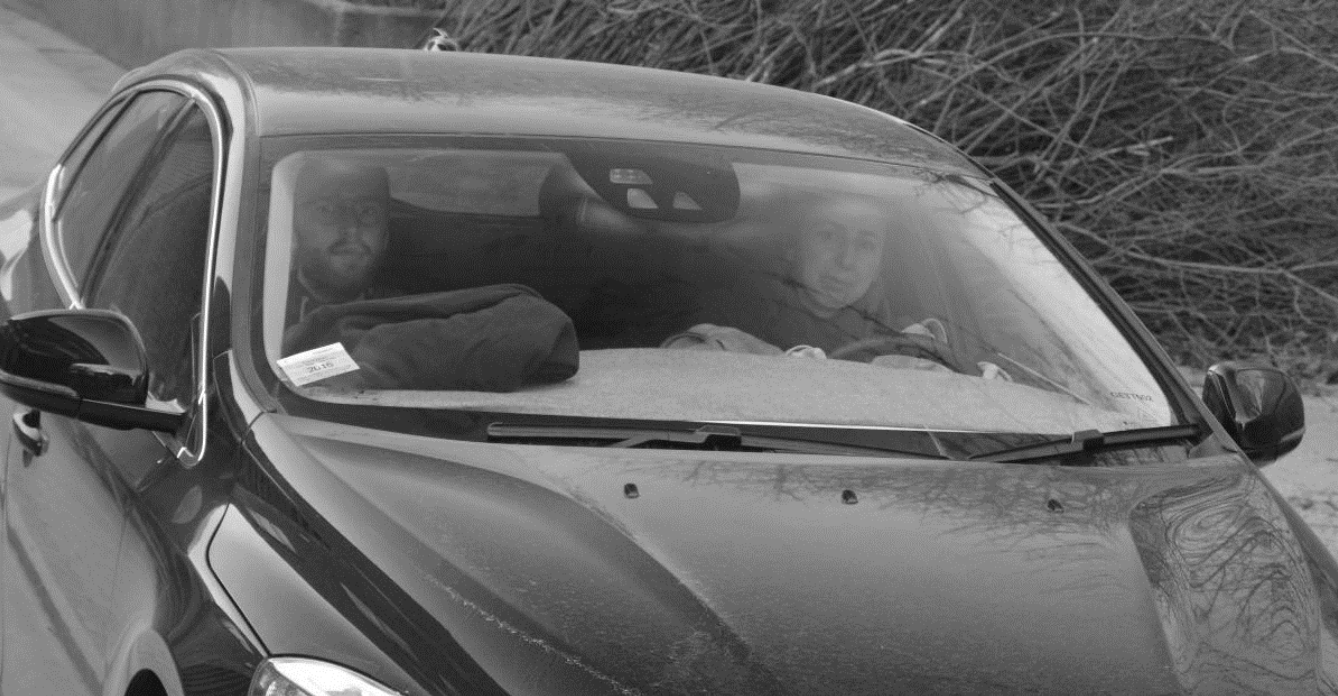}
\label{fig:pretend_driver} }
\caption[wizard of oz examples]{Examples of Wizard of Oz techniques to disguise the driver. a) The driver is disguised as a car seat \cite{dey2017impact} and b) the driver is driving the car from a right-hand steering wheel while a dummy driver is sitting in the actual driver's seat \cite{lagstrom2015avip}.}
\vspace*{-0.5cm} 
\label{fig:wizard_of_oz}
\end{figure}

In the context of autonomous driving research, the Wizard of Oz technique \cite{lagstrom2015avip} is common in which the experimenters simulate the behavior of an intelligent system to observe the reaction of subjects. Using this technique, experimenters may disguise themselves as a car seat \cite{dey2017impact} or control the vehicle from a hidden place inside the vehicle \cite{lagstrom2015avip} that is not observable by the participants (see \figref{fig:wizard_of_oz}). 

\begin{figure}[!t]
\centering
\includegraphics[width=\columnwidth]{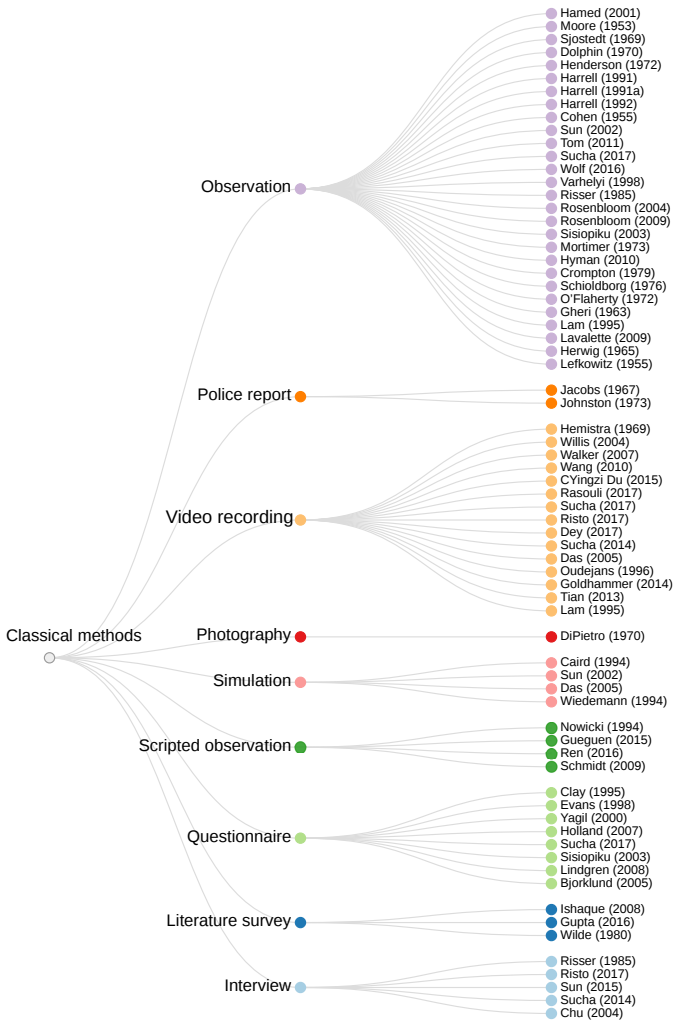}
\caption[factors and papers.]{Data collection methods used in the classical pedestrian behavior studies.}
\vspace*{-0.3cm} 
\label{fig:ped_studies_methods}
\end{figure}

\begin{figure}[!t]
\centering
\includegraphics[width=\columnwidth]{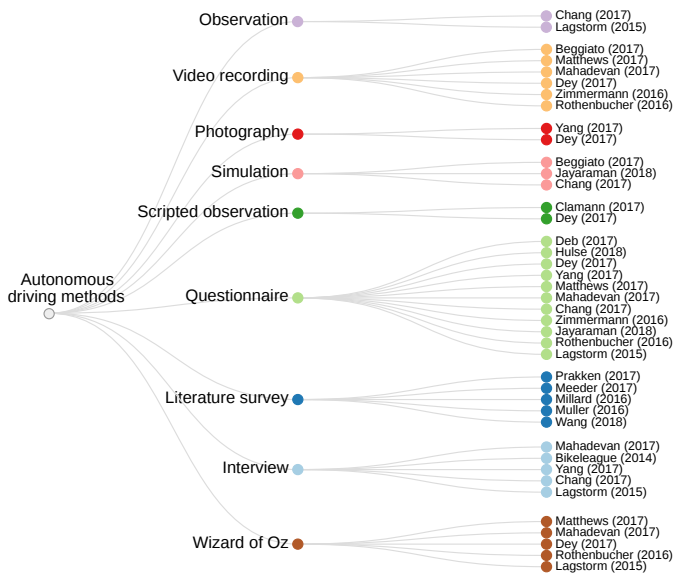}
\caption[factors and papers.]{Data collection methods used in the pedestrian behavior studies involving autonomous vehicles.}
\vspace*{-0.5cm} 
\label{fig:ped_studies_auto_methods}
\end{figure}

Figures \ref{fig:ped_studies_methods} and \ref{fig:ped_studies_auto_methods} summarize the works presented in this paper and their methods of study. Note that in this figure literature survey refers to expert studies that generate new findings based on past works.

\section{Pedestrian Behavior Studies}

We divide  pedestrian behavior studies into two categories, classical studies and  ones involving autonomous vehicles. Compared to studies with autonomous vehicles, the classical studies focus on pedestrian behavior while interacting with human drivers instead of vehicles. All the factors identified in the literature are italicized in the text. 

\subsection{Classical Studies}
\label{classical_studies}

The early works in pedestrian behavior studies come from early 1950s, and since then there has been a tremendous amount of research done on various factors that impact pedestrian behavior. Given the magnitude of the work in this area, an exhaustive survey of all the literature would be prohibitive. As a result, only a subset of major works will be presented.

\begin{figure*}[th]
\centering
\includegraphics[width=\textwidth]{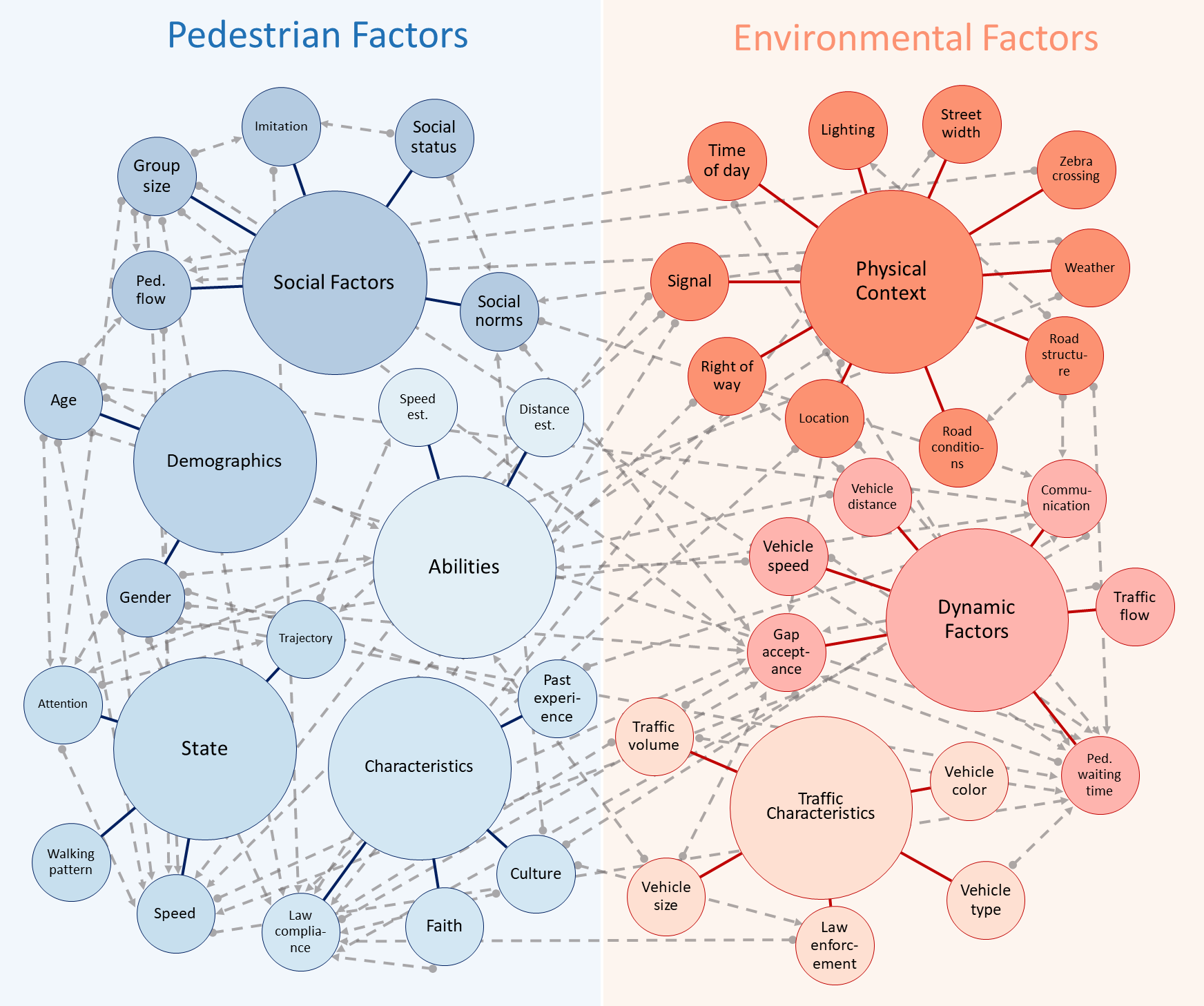}
\caption[Factors involved in pedestrian crossing behavior.]{Factors involved in pedestrian decision-making process at the time of crossing. The circles refer to the factors, the branches with solid lines indicate the sub-factors of each category and the dashed lines show the interconnection between different factors and arrows show the direction of influence.}
\vspace*{-0.5cm} 
\label{fig:ped_factors_relationship}
\end{figure*}

We divide the factors that influence pedestrian behavior into two groups, the ones that directly relate to pedestrians (e.g. demographics) and environmental ones (e.g. traffic conditions). A summary of these factors and how they are interrelated can be found in \figref{fig:ped_factors_relationship}.

\subsubsection{Pedestrian Factors}

\textbf{\textit{Social Factors}}. Among the social factors, perhaps, \textit{group size} is one of the most influential ones. Heimstra \etal \cite{heimstra1969experimental} conducted a naturalistic study to examine the crossing behavior of children and found that they commonly (in more than 80\% of the cases) tend to cross as a group rather than individually. \textit{Group size} changes both the behavior of the drivers with respect to the pedestrians and the way the pedestrians act at crosswalks. For instance, it is shown that drivers more likely yield to groups of pedestrians (3 or more) than individuals \cite{sun2002modeling,herwig1965verhalten}.

When crossing as a group, pedestrians tend to be more careless, and pay less \textit{attention} at crosswalks and often accept shorter gaps between the vehicles to cross \cite{wang2010study,schioldborg1976children,harrell1991factors} or do not look for approaching traffic \cite{sucha2017pedestrian}. \textit{Group size} is also found to impact the way pedestrians comply with the traffic laws, i.e. \textit{group size} exerts some form of social control over individual pedestrians \cite{rosenbloom2009crossing}. It is observed that individuals in a group are less likely to follow a person who is breaking the law, e.g. crossing on the red light \cite{lefkowitz1955status}.

In addition, \textit{group size}, for obvious reasons, influences \textit{pedestrian flow} which determines how fast pedestrians cross the street. Wiedemann \cite{wiedemann1994simulation} indicates that if there is no interaction between the pedestrians, there is a linear relationship between \textit{pedestrian flow} and \textit{pedestrian speed}. This means, in general, pedestrians walk slower in denser groups.

\textit{Social norms}, or as some experts refer to as "informal rules" \cite{farber2016communication}, play a significant role in how traffic participants behave and how they predict each other's intention \cite{wilde1980immediate}. \textit{Social norms} also influence how acceptable a particular action is in a given traffic situation \cite{evans1998understanding}. The difference between \textit{social norms} and legal norms (or formal rules) can be illustrated using the following example: formal rules define the speed limit of a street, however, if the majority of drivers exceed this limit, the \textit{social norm} is then quite different \cite{wilde1980immediate}.

The influence of \textit{social norms} is so significant that merely relying on formal rules does not guarantee safe interaction between traffic participants. This fact is highlighted in a study by Johnston \cite{johnston1973road} in which he describes the case of a 34-year old married woman who was extremely cautious (and often hesitant) when facing yield and stop signs. In a period of four years, this driver was involved in 4 accidents, none of which she was legally at fault. In three out of four cases the driver was hit from behind, once by a police car. This example illustrates how disobeying \textit{social norms}, even if it is legal,  can disrupt traffic flow.

\textit{Social norms} even influence the way people interpret the law. For example, the concept of "psychological right of way" or "natural right of way" has been studied \cite{wilde1980immediate}. This concept describes the situation in which drivers want to cross a non-signalized intersection. The law requires the drivers to yield to the traffic from the right. However, in practice drivers may do quite the opposite depending on the \textit{social status} (or configuration) of the street. It is found that factors such as \textit{street width}, \textit{lighting} conditions or the presence of shops may determine how the drivers would behave \cite{gheri1963blickverhalten}.

\textit{Imitation} is another social factor that defines the way pedestrians (as well as drivers \cite{vsucha2014road}) would behave. A study by Yagil \cite{yagil2000beliefs} shows that the presence of a law-adhering (or law-violating) pedestrian increases the likelihood of other pedestrians to obey (or disobey) the law. This study shows that the impact is more significant when law violation is involved.

The probability of \textit{imitation} occurrence may depend on the \textit{social status} of the person who is being imitated. In the study by Leftkowitz \etal \cite{lefkowitz1955status} a confederate was asked by the experimenter to cross or stand on the sidewalk. The authors observed that when the research confederate was wearing a fancy outfit, there was a higher chance that other pedestrians imitate his actions (either breaking the law or complying). This idea, however, is challenged by Dolphin \etal \cite{dolphin1970factors} whose findings indicate that \textit{social status} and \textit{gender} have no effect on \textit{imitation}. The authors claim that \textit{group size} is a better predictor for the likelihood of \textit{imitation}, which means the larger the size of the group, the lower the chance of pedestrians imitating others.  

\textbf{\textit{Demographics}}. Arguably, \textit{gender} is one of factors that influences pedestrian behavior the most \cite{heimstra1969experimental,holland2007effect,moore1953pedestrian}. Studies show that women in general are more cautious than men \cite{heimstra1969experimental,holland2007effect,yagil2000beliefs} and demonstrate a higher degree of \textit{law compliance} \cite{tom2011gender,jacobs1967study}.

Furthermore, \textit{gender} differences affect the motives of pedestrians when complying with the law. Yagil \cite{yagil2000beliefs} argues that crossing behavior in men is mainly predicted by normative motives (the sense of obligation to the law) whereas in women it is better predicted by instrumental motives (the perceived danger or risk). He adds that women are influenced by social values, e.g. what other people think about them, while men are mainly concerned with physical conditions, e.g. the structure of the street.

Men and women also differ in the way they pay \textit{attention} to the environment before or during crossing. For instance, Tom and Granie \cite{tom2011gender} show that prior to and during a crossing event, men more frequently look at vehicles whereas women look at traffic lights and other pedestrians, i.e. they have different \textit{attention} patterns. Women also tend to change their gazing pattern according to \textit{road structure}, show a higher behavior variability \cite{holland2007effect}, and cross with a lower \textit{speed} compared to men \cite{ishaque2008behavioural}.

\textit{Age} impacts pedestrian behavior in obvious ways. Generally, elderly pedestrians are physically less capable compared to adults, and as a result, they walk slower \cite{ishaque2008behavioural}, have a more varied \textit{walking pattern} (e.g. do not have steady velocity) \cite{goldhammer2014analysis} and are more cautious in terms of \textit{gap acceptance}  \cite{sun2002modeling,harrell1991precautionary}. Being more cautious  means older pedestrians, compared to adults and children,  spend a longer time paying \textit{attention} to the traffic prior to crossing \cite{rasouliagree}. Furthermore, the elderly and children are found to have a lesser ability to correctly assess the speed of vehicles, hence are more vulnerable \cite{clay1995driver}. It is also interesting to note that there is a higher variability observed in younger pedestrians' behavior, making them less predictable \cite{holland2007effect}.  

\textbf{\textit{State}}. The \textit{speed} of pedestrians is thought to influence their visual perception of dynamic objects. Oudejans \etal \cite{oudejans1996cross} argue that while walking, pedestrians have better optical flow information, and consequently, a better sense of \textit{speed and distance estimation}. Thus walking pedestrians are less conservative to cross compared to standing ones.

Pedestrian \textit{speed} may vary depending on the conditions such as \textit{road structure}. For instance, pedestrians tend to walk faster during crossing compared to when they walk on sidewalks \cite{tian2013pilot} or walk faster on wider sidewalks as the density of pedestrians can be lower \cite{moore1953pedestrian}. When vehicles have \textit{the right of way} or  pedestrians' \textit{trajectory} is towards the vehicles, they tend to cross faster \cite{tian2013pilot}. In addition, \textit{road structure} impacts crossing \textit{speed}. For example, Crompton \cite{crompton1979pedestrian} reports pedestrian mean speed at different crosswalks as follows: 1.49 m/s at \textit{zebra crossings}, 1.71 m/s as crossing with pedestrian refuge island and 1.74 m/s at pelican crossings.

Other factors that have been shown to affect pedestrian \textit{speed} include \textit{group size}, generally slower in larger groups, \cite{dipietro1970pedestrian,O1972movement,willis2004human}, \textit{age}, pedestrians tend to get slower as they age, \cite{sjostedt1969behaviour,willis2004human}, \textit{time of day}, generally walk faster in early morning rush, and \textit{road structure}, if there is more space for pedestrians, they tend to walk faster \cite{willis2004human}.

The effect of \textit{attention} on traffic safety has been extensively studied in the context of driving \cite{underwood2003visual,klauer2005driver,underwood2007visual,barnard2016study}. As for pedestrians, it is shown that the majority of pedestrians tend to pay \textit{attention} prior to crossing, the frequency of which may vary depending on the crosswalk delineation such as the presence of traffic \textit{signals} or \textit{zebra crossing} lines \cite{rasouliagree}. Some findings suggest that when pedestrians make eye contact with drivers, the drivers are more likely to slow down and yield to the pedestrians \cite{ren2016analysis}.

Hymann \etal \cite{hyman2010did} investigate the effect of \textit{attention} on pedestrian walking \textit{trajectory}. They show that pedestrians who are distracted by the use of electronics, such as mobile phones, are 75\% more likely to display inattentional blindness (not noticing the elements in the scene). Distracted pedestrians often change their walking direction and, on average, walk slower than undistracted pedestrians.

\textit{Trajectory} or pedestrian walking direction is another factor that plays a role in the way pedestrians make a crossing decision. Schmidt and Farber \cite{schmidt2009pedestrians} argue that when pedestrians are walking in the same direction as the vehicles, they tend to make riskier decisions regarding whether to cross. According to the authors, walking direction can alter the ability of pedestrians to estimate speed. In fact, pedestrians have a more accurate \textit{speed estimation} when the approaching cars are coming from the opposite direction.

\textbf{\textit{Characteristics}}. Among different pedestrian characteristics,  \textit{culture} plays an important role. It defines the way people think and behave, and forms a common set of \textit{social norms} they obey  \cite{lindgren2008requirements}. Variations in traffic \textit{culture}  exist not only between different countries  but also within the same country, e.g. between towns and countrysides or between different cities \cite{bjorklund2005driver}.

A number of studies connect \textit{culture} to the types of behavior that road users exhibit. Lindgren \etal \cite{lindgren2008requirements} compare the behaviors of Swedish and Chinese drivers and show that they assign different levels of importance to various traffic problems such as speeding or jaywalking. Schmidt and Farber \cite{schmidt2009pedestrians} point out the differences in \textit{gap acceptance} of Indians (2-8s) versus Germans (2-7s). Clay \cite{clay1995driver} indicates the way people from different culture perceive and analyze a situation. She notes that Americans judge traffic behavior based on characteristics of the pedestrians whereas Indians rely more on contextual factors such as traffic condition, road structure, etc.

Some researchers go beyond \textit{culture} and study the effect of \textit{faith} or religious beliefs on pedestrian behavior. Rosenbloom \etal \cite{rosenbloom2004heaven} gather that ultra-orthodox (in a religious sense) pedestrians in an ultra-orthodox setting are three times more likely to violate traffic laws than secular pedestrians.

Generally speaking, pedestrian level of \textit{law compliance} defines how likely they would break the law (e.g. crossing at red light). In addition to demographics, \textit{law compliance} can be influenced by physical factors, for instance, the \textit{location} of a designated crosswalk influences the decision of pedestrians whether to jaywalk \cite{sisiopiku2003pedestrian}. 

Another factor that characterizes a pedestrian is his/her \textit{past experience}. For example, non-driver female pedestrians generally tend to be more cautious when making crossing decision \cite{holland2007effect}. 

\textbf{\textit{Abilities}}. The ability to \textit{estimate speed and distance}, can influence the way pedestrians perceive the environment and consequently the way they react to it. In general, pedestrians are better at judging \textit{vehicle distance} than \textit{vehicle speed} \cite{sun2015estimation}. For instance, they can correctly estimate \textit{vehicle speed} when the vehicle is moving below the speed of 45 km/h, whereas \textit{vehicle distance} can be correctly estimated when the vehicle is moving up to a speed of 65 km/h.

\subsubsection{Environmental Factors}

\textbf{\textit{Physical context}}. The presence of street delineations, including traffic \textit{signals} or \textit{zebra crossings}, has a major effect on the way traffic participants behave \cite{moore1953pedestrian}, or on their degree of \textit{law compliance} \cite{mortimer1973behavioral}. Some scholars distinguish between the way traffic \textit{signals} and \textit{zebra crossings} influence yielding behavior. For example, traffic signals (e.g. traffic light) prohibit vehicles to go further and force them to yield to crossing pedestrians. At non-signalized \textit{zebra crossings}, however, drivers usually yield if there are pedestrians present at the curb who either clearly communicate their intention of crossing (often by eye contact) or start crossing (by stepping on the road) \cite{sucha2017pedestrian}.

\textit{Signals} alter pedestrians level of cautiousness as well \cite{rasouliagree}. In a study by Tom and Granie \cite{tom2011gender}, it is shown that pedestrians look at vehicles 69.5\% of the time at signalized and 86\% of the time at unsignalized intersections. In addition, the authors point out that pedestrians' \textit{trajectory} differs at unsignalized crossings, i.e. they tend to cross diagonally when no signal is present. 

Some studies discuss the likelihood of pedestrians to use dedicated \textit{zebra crossing}. In general, women and children use dedicated zebra crossings more often \cite{moore1953pedestrian,jacobs1967study}. \textit{Traffic volume} and the presence of \textit{law enforcement} personnel near crossing lines are also shown to induce pedestrians to use designated crossing lines. The effect of \textit{law enforcement}, however, is much stronger on men than women \cite{moore1953pedestrian}. 

In terms of crossing \textit{speed}, pedestrians tend to walk faster at signalized crosswalks \cite{lam1995pedestrian,mortimer1973behavioral}. The presence of signals also induces pedestrians to comply with the law, although this effect seems to be opposite for one-way streets \cite{de2009pedestrian}. 

\textit{Road structure} (e.g. crossing type and road geometry) and \textit{street width} impact the level of crossing risk (or affordance) \cite{oudejans1996cross}. For example, pedestrians  pay more \textit{attention} prior to crossing in wide streets \cite{rasouliagree} and accept a smaller gap in narrow streets \cite{schmidt2009pedestrians,rasouliagree}. \textit{Road structure} is also believed to alter the way drivers behave, which subsequently can influence pedestrians' expectations \cite{bjorklund2005driver}.

With respect to \textit{law compliance}, contradictory findings have been reported. While some researchers claim larger \textit{street width} can increase the chance of compliance \cite{chu2004people}, others report the opposite and show it can increase crossing violation \cite{de2009pedestrian}.

\textit{Weather} or \textit{lighting} conditions affect pedestrian behavior in many ways \cite{harrell1991factors}. For instance, in bad \textit{weather} conditions pedestrians' \textit{speed estimation} is poor, therefore they become conservative while crossing \cite{sun2015estimation}. Pedestrians (especially the elderly and women) are found to be more cautious in warm weather than cold \cite{harrell1991factors}. Moreover, lower illumination level (e.g. nighttime) reduces pedestrians' major visual functions (e.g. resolution acuity, contrast sensitivity and depth perception), causing them to make riskier decisions. Another direct effect of \textit{weather} would be on \textit{road conditions}, such as slippery roads due to rain, that can impact movements of both drivers and pedestrians \cite{lin2016impact,moore1953pedestrian}.

\textbf{\textit{Dynamic factors}}. One of the key dynamic factors is \textit{gap acceptance} or how much gap in traffic   (typically in time) pedestrians consider safe to cross. \textit{Gap acceptance} depends on two dynamic factors, \textit{vehicle speed} and \textit{vehicle distance} from the pedestrian. The combination of these two factors defines Time To Collision (or Contact) (TTC), or how far the approaching vehicle is from the point of impact \cite{yangpedestrian,das2005walk,rasouliagree}. The average pedestrian \textit{gap acceptance} is between 3-7s, i.e. usually pedestrians do not cross when TTC is below 3s \cite{dipietro1970pedestrian} and very likely cross when it is higher than 7s \cite{schmidt2009pedestrians}. As mentioned earlier, \textit{gap acceptance} may highly vary depending on social factors (e.g. \textit{demographics} \cite{wang2010study,harrell1992gap}, \textit{group size} \cite{dipietro1970pedestrian}, \textit{culture} \cite{schmidt2009pedestrians}), level of \textit{law compliance} \cite{ishaque2008behavioural}, and the \textit{street width}. For instance, women and the elderly generally accept longer gaps \cite{cohen1955risk} and people in groups accept a shorter time gap \cite{harrell1992gap}.  

The effects of \textit{vehicle speed} and \textit{vehicle distance} are also studied in isolation. It is shown that increase in \textit{vehicle speed} deteriorates pedestrians' ability to estimate speed \cite{clay1995driver} and distance \cite{sun2015estimation}. In addition, pedestrians are found to rely more on distance when crossing, i.e. within the same TTC, and they cross more often when the speed of the approaching vehicle is higher \cite{schmidt2009pedestrians}.

Some scholars look at the relationship between pedestrian \textit{waiting time} prior to crossing and \textit{gap acceptance}. Sun \etal \cite{sun2002modeling} argue that the longer pedestrians wait, the more frustrated they become and, as a result, their   \textit{gap acceptance} lowers. The impact of \textit{waiting time} on crossing behavior, however, is controversial. Wang \etal \cite{wang2010study} dispute the role of \textit{waiting time} and claim that in isolation \textit{waiting time} does not explain the changes in \textit{gap acceptance}. They add that to be considered effective, \textit{waiting time} should be studied in conjunction with other factors such as pedestrians' personal characteristics.

Pedestrian \textit{waiting time} can be influenced by a number of factors such as \textit{age}, \textit{gender}, \textit{road structure}, \textit{location} (e.g. how close to one's destination) and pedestrian \textit{walking speed}. Females are generally  have longer \textit{waiting time} compared to men \cite{dipietro1970pedestrian,hamed2001analysis}. Pedestrians who can walk faster (which is affected also by \textit{age}) tend to spend less time waiting prior to crossing \cite{hamed2001analysis}. In terms of \textit{road structure}, studies show that, when crossing a road with a refuge island, pedestrians cross faster from one side to the island than the island to the other side.

Although \textit{traffic flow} is a byproduct of \textit{vehicle speed and distance}, on its own it can also be a predictor of pedestrian crossing behavior \cite{schmidt2009pedestrians}. By observing the overall pattern of traffic, pedestrians might form an expectation about what approaching vehicles might do next.

The role of \textit{communication} (often nonverbal) in resolving traffic ambiguities is emphasized by a number of scholars \cite{wilde1980immediate,clay1995driver,sucha2017pedestrian}. In this context, any kind of signal between road users constitutes communication. In traffic scenes, communication is particularly precarious because, firstly, there exists no official set of signals and most of them are ambiguous, and secondly, the type of communication may change depending on the atmosphere of the traffic situation, e.g. city or country \cite{risser1985behavior}.

The lack of \textit{communication} or miscommunication can greatly contribute to traffic conflicts. It is shown that more than a quarter of traffic conflicts is due to the absence of effective communication between road users. In particular, pedestrians heavily rely on communication when making crossing decisions and report feeling uncomfortable when the communication is non-existent and certain vehicle behaviors are not observed \cite{risto2017human}.  

Traffic participants use different methods to \textit{communicate} with each other. For example, pedestrians use eye contact (gazing/staring), a subtle movement in the direction of the road, handwave, smile or head wag. Drivers, on the other hand, flash lights, wave hands or make eye contact \cite{sucha2017pedestrian}. Some researchers also point out that the speed changes of the vehicle can be an indicator of the driver's intention \cite{rasouliagree}. For example, in a case study by Varhelyi \cite{varhelyi1998drivers} it is shown that drivers maintain their speed or accelerate to communicate their intention of not yielding to pedestrians. This means pedestrian reaction (or intention of crossing) may vary depending on the behavior of drivers. The stopping behavior of vehicles may also contain a communicational cue. Studies show when drivers stop their cars far shorter than where they legally must stop, they are signaling their intention of giving the \textit{right of way} to others \cite{dey2017pedestrian}.

Among different forms of nonverbal \textit{communication}, eye contact is particularly important. Pedestrians often establish eye contact with drivers to make sure they are seen \cite{rasouli2017understanding}. Drivers also often make eye contact and gaze at the face of other road users to assess their intentions \cite{walker2007drivers}. It is found that the presence of eye contact between road users increases compliance with instructions and rules \cite{gueguen2015pedestrian}. For instance, drivers who make eye contact with pedestrians will more likely yield right of way at crosswalks \cite{gueguen2015pedestrian}.

According to a study by Dey \etal \cite{dey2017pedestrian}, the majority of communication in traffic is implicit (e.g. walking behavior) rather than explicit (e.g. hand gestures). They report that nearly 97\% of pedestrians do not engage in any form of explicit communication with drivers. About 63\% of pedestrians claim their \textit{right of way} simply by stepping on the road. 

The authors of \cite{dey2017pedestrian} argue that pedestrians treat vehicles as entities and do not care about the state of the driver when making crossing decision. Even though at the time of crossing pedestrians look towards the approaching vehicles, they do not engage in eye contact and rather observe the state of the vehicle. These findings, however, are questionable. Overall, there is much stronger support for the role of eye contact in crossing actions (refer to \textit{attention}), with the authors themselves admitting that during their study they had no way of accurately tracking pedestrians' (or drivers') gaze.  

When speaking of \textit{communication}, two additional factors should be considered, namely \textit{culture} and \textit{social norms} which determine the type and the meaning of communication signals used by road users \cite{wolf2016interaction}. For example, Gupta \etal \cite{gupta2016conventionalized} show how in Germany raising one hand by a police officer means the attention command, whereas in India the same command is communicated by raising both hands. 

\begin{figure*}[!t]
\centering
\includegraphics[width=\textwidth]{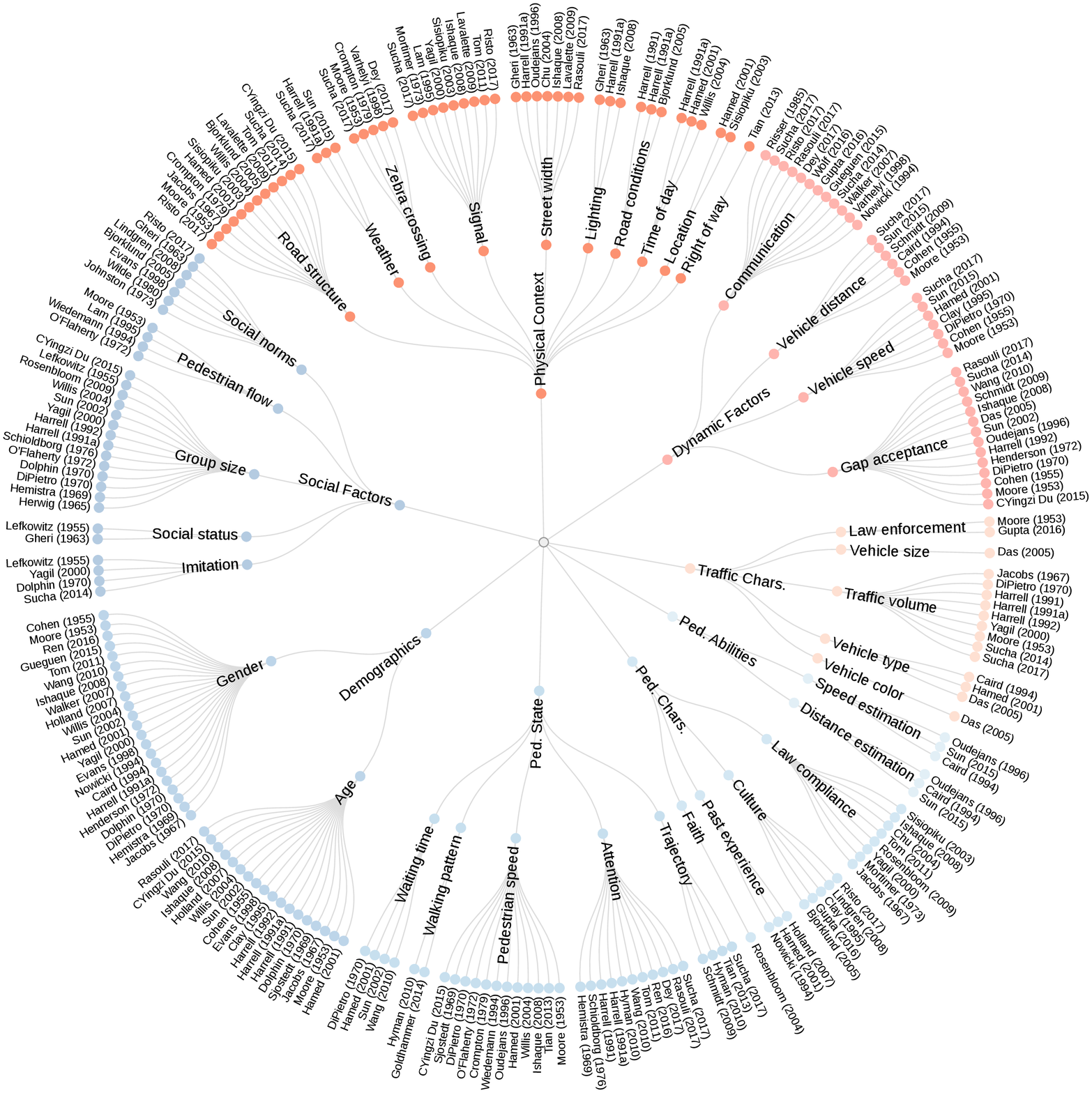}
\caption[factors and papers.]{A circular dendrogram of the factors influencing pedestrian behavior and the classical studies that identified them. Leaf nodes represent the individual studies (identified by the first author and year of publication) and internal nodes represent minor and major factors.}
\vspace*{-0.5cm} 
\label{fig:ped_factors_all}
\end{figure*}

\textbf{\textit{Traffic characteristics}}. \textit{Traffic volume} or density affects pedestrian \cite{vsucha2014road} and driver behavior \cite{schmidt2009pedestrians} significantly. Essentially, the higher the density of traffic, the lower the chance of pedestrians to cross \cite{ishaque2008behavioural}. This is particularly true when it comes to \textit{law compliance}, i.e. pedestrians are less likely to cross against the \textit{signal} (e.g. red light) if the traffic volume is high. The effect of \textit{traffic volume}, however, is stronger on male pedestrians than women \cite{yagil2000beliefs}.

The effects of vehicle characteristics such as \textit{vehicle size} and \textit{vehicle color} on pedestrian behavior have been investigated. Although \textit{vehicle color} has not shown to have a measurable effect, \textit{vehicle size} can influence crossing behavior in two ways. First, pedestrians tend to be more cautious when facing a larger vehicle \cite{das2005walk}. Second, the size of the vehicle impacts pedestrian \textit{speed and distance estimation} abilities. In an experiment involving 48 men and women, Caird and Hancock \cite{caird1994perception} reveal that as the size of the vehicle increases, there is a higher chance that people will underestimate its arrival time.

When making a crossing decision, the \textit{vehicle type} matters and can influence different \textit{genders} differently. For example, compared to women, men are generally better in judging the type of vehicles and are more accurate at estimating the arrival time of vans and motorcycles \cite{caird1994perception}. In addition, pedestrians exhibit different \textit{waiting time} when facing different types of vehicles, e.g. they tend to cross faster in front of passenger vehicles \cite{hamed2001analysis}.  

A summary of the factors from the classical literature is illustrated in \figref{fig:ped_factors_all}. Here we can see that more studies have been conducted on factors such as \textit{gender}, \textit{group size}, \textit{age} and  \textit{gap acceptance}, compared to \textit{culture}, \textit{vehicle size}, \textit{right of way}, and \textit{faith}. Due to the emergence of intelligent transportation systems and the availability of technology for collecting data, studies on factors such as \textit{communication}, \textit{attention}, \textit{pedestrian trajectory} and \textit{culture} have gained popularity in the past few years. However, a number of factors such as \textit{lighting}, \textit{road conditions}, \textit{vehicle type}, \textit{past experience}, \textit{social status}, and \textit{pedestrian flow} are left unaddressed in recent works.

\subsection{Studies in the Context of Autonomous Driving}
\label{beh_autonomous}

\begin{figure*}[!ht]
\centering
\includegraphics[width=0.9\textwidth]{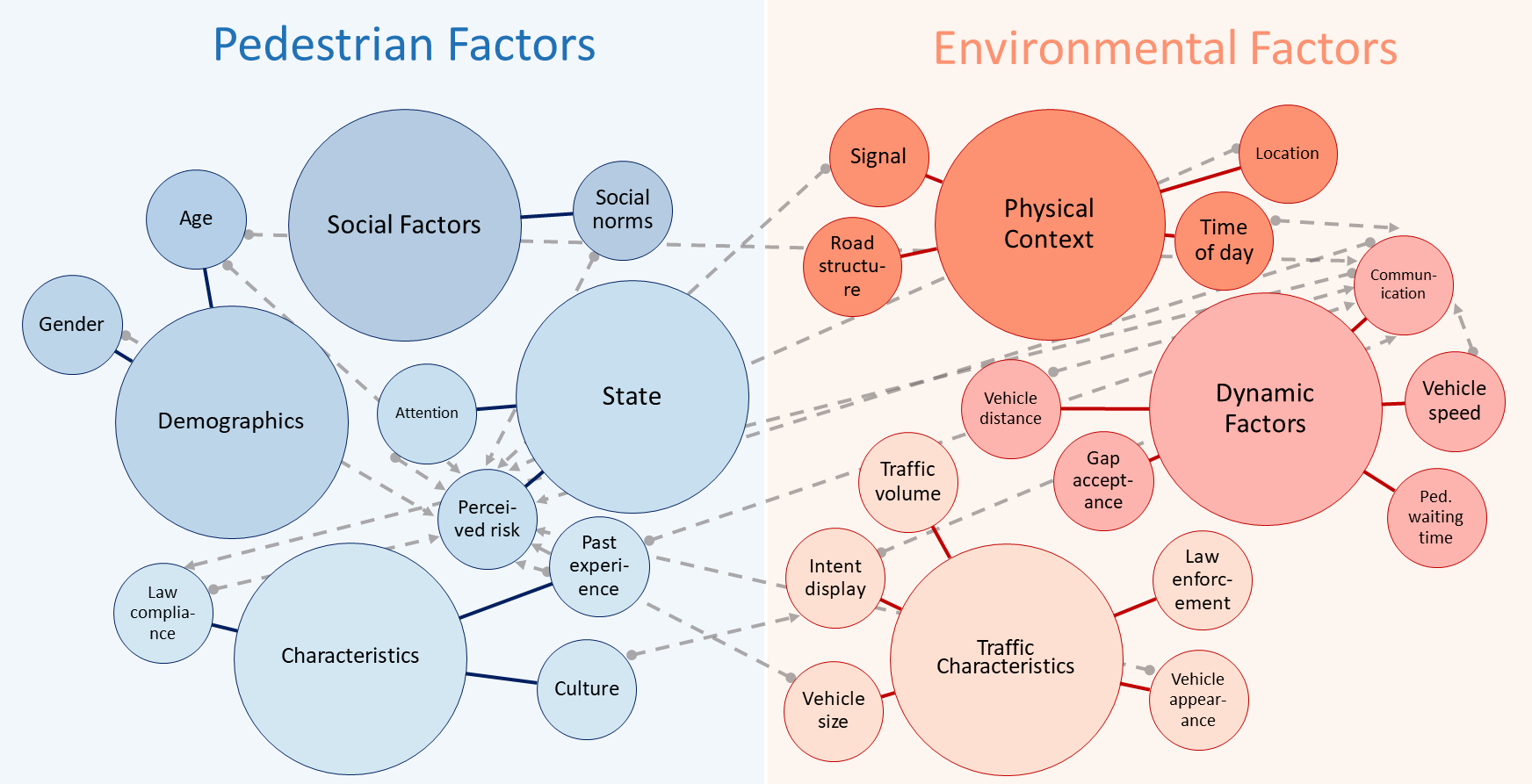}
\caption[Factors involved in pedestrian crossing behavior.]{Factors involved in pedestrian decision-making process when facing autonomous vehicles. The circles refer to the factors, the branches with solid lines indicate the sub-factors of each category and the dashed lines show the interconnection between different factors and arrows show the direction of influence.}
\vspace*{-0.4cm} 
\label{fig:ped_factors_connection_auto}
\end{figure*}

Similar to classical studies, we divide behavioral studies involving autonomous vehicles into two groups of pedestrian and environmental factors. A summary of these factors and their connections can be found in \figref{fig:ped_factors_connection_auto}.

\begin{figure}[!tp]
\centering
\captionsetup[subfigure]{labelformat=empty}

\includegraphics[width=0.30\textwidth]{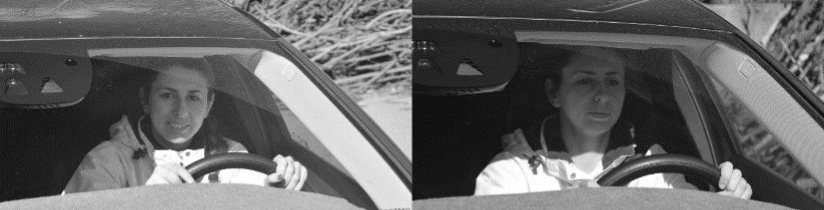}\hspace{-0.125cm}
\includegraphics[width=0.15\textwidth]{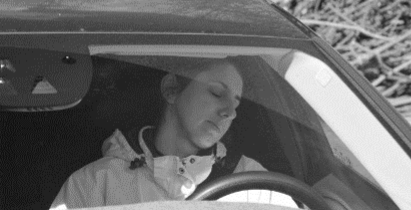}\\
\includegraphics[width=0.30\textwidth]{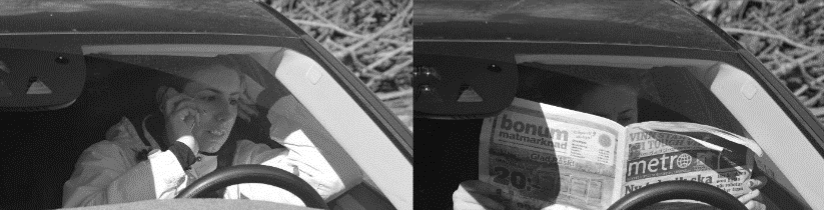}\hspace{-0.125cm}
\caption[driver's conditions]{Driver's conditions used in the experiments conducted in \cite{lagstrom2015avip}.}
\vspace*{-0.5cm} 
\label{fig:driver_conditions}
\end{figure}

Studies concerning the social aspects of autonomous driving generally focus on two major factors, namely \textit{communication} and \textit{attention}. Regarding the necessity of \textit{communication}, the autonomous driving community is divided. Millard \cite{millard2016pedestrians} argues that the interaction between pedestrians and autonomous vehicles resembles, what he refers to as the game of "crosswalk chicken". In a normal situation involving a human driver, if a pedestrian chooses to cross, they accept a large risk because, the norms permits not yielding to pedestrians, the driver might be distracted or assume the pedestrian would not intend to cross. According to Millard, in the case of autonomous driving the \textit{perceived risk} of crossing is almost nonexistent because the pedestrian knows that the autonomous vehicle will stop, and as a result there is no need for any form of \textit{communication} to reach an agreement with the vehicle. Using field studies, Rothenbucher \etal \cite{rothenbucher2016ghost} support the same argument and show that without \textit{communication} and \textit{attention} (the need for establishing eye contact), when facing an autonomous vehicle, pedestrians eventually adjust their behavior and cross the street. The result of this study, however, is questionable because the trials took place on a university campus where the speed limit was very low and the vehicle posed minimal threat to pedestrians. The subjects who were observed or participated in the interviews may also have heard about the experiment, or in general, had higher acceptance compared to general population for autonomous driving technologies.

Overall, arguments in favor of \textit{communication} necessity in autonomous driving are stronger. A number of studies relate to existing literature and \textit{past experience} to support the role of \textit{communication} \cite{muller2016social,meeder2017autonomous,prakken2017problem,wang2018act}. Muller \cite{muller2016social} argues that identifying autonomous vehicles in traffic is not always intuitive. Road users might recognize an autonomous vehicle as a traditional vehicle and expect certain behaviors from it. As for the need for \textit{communication}, the author describes a busy pedestrian crossing where a driver might communicate his intention by moving forward slowly into the crowd. The author then raises concern regarding how an autonomous vehicle would behave in such a situation. 

The \textit{communication} necessity can also be seen from a different perspective. Prakken \cite{prakken2017problem} emphasizes the importance of understanding communicational cues in obeying traffic laws. He mentions that the current technology does not distinguish between the type of pedestrians which can be problematic when a \textit{law enforcement} officer is present in the scene for directing the traffic. According to Prakken autonomous vehicles should be able to interpret and distinguish communication messages produced by \textit{law enforcement} personnel and regular pedestrians.

There are a number of empirical studies that support the role of \textit{communication} and \textit{attention} in autonomous driving. A survey conducted by the League of American Bicyclists \cite{bikeleauge2014} shows that besides issues related to technological advancements, inability to communicate and establishing eye contact are among major reasons that increase pedestrians and bicyclists \textit{perceived risk} when interacting with autonomous vehicles.

Lagstrom and Lundgren \cite{lagstrom2015avip}, and, in a later study, Yang  \cite{yang2017driver} evaluate the role of driver behavior when the vehicle is running autonomously. The authors used several scenarios of driver behavior when crossing an intersection including the driver making eye contact, staring straight at the front road, talking on the phone, reading a newspaper and sleeping (see \figref{fig:driver_conditions}). In these experiments, the vehicles were operated by drivers (who were hidden from the view of pedestrians) using a right-hand steering wheel. Observing pedestrians' reactions, Lagstrom and Lundgren show that when the vehicle was stopping and the driver paid \textit{attention} (made eye contact) to pedestrians, all pedestrians crossed the street. However, when the driver was busy on the phone, 20\% of pedestrians did not cross and when the driver was reading a newspaper or not present in the vehicle, 60\% of the pedestrians did not cross. In both studies surveys were conducted to measure the pedestrians' level of \textit{perceived risk} in each situation. The results show that when a form of \textit{attention} (eye contact) was present, the pedestrians felt most comfortable. Yang \cite{yang2017driver} also adds that \textit{vehicle appearance} impacts the level of pedestrians' comfort. Her findings indicate that when the pedestrians could not see the driver (due to dark windows), they felt most uncomfortable.

Matthews \etal \cite{matthews2017intent} measure the importance of using an \textit{intent display} in \textit{communication} with pedestrians. The authors used a remotely controlled golf cart with and without an intent display mechanism. They observed that when the vehicle equipped with a display was encountering pedestrians, there was 38\% improvement in resolving deadlocks. The authors show that the improvement can increase based on the pedestrians' \textit{past experience}. The group of participants who were familiarized with the communication technology prior to the experiment exhibited more trust in the vehicle.

Although \textit{intent displays} have been shown to improve the overall experience of pedestrians during interaction \cite{matthews2017intent,zimmermann2017first}, they don't always seem to be very effective. In her studies, Yang \cite{yang2017driver} used a display to show "Safe to Cross" message to pedestrians. When interviewed by the experimenter, the participants responded that the display did not have a significant effect on their crossing decision. In another study, Clamann \etal \cite{clamann2017evaluation} found that pedestrians still focus on legacy factors such as \textit{vehicle speed and distance} when making crossing decision. The use of the display only influenced 12\% of the participants' decisions and overall increased the time of decision-making. In this context, however, the authors show that informative displays (e.g. with information about vehicle's speed) compared to advisory displays (e.g. cross or not to cross signal) are more effective. The authors add that the traditional social and environmental factors such as \textit{age}, \textit{gender} \textit{road structure}, \textit{waiting time} and \textit{traffic volume} are still very important in the context of autonomous driving.

Other forms of intention \textit{communication} methods have also been examined. Chang \etal \cite{chang2017eyes} propose the use of moving eyes installed at the front of the vehicles. Using experimental data collected from 15 participants, the authors show that more than 66\% of participants made street crossing decision faster in the presence of eyes, and if the eyes were looking at the participants, this number rose to more than 86\%. The empirical evaluation of this study, however, is limited to virtual reality environment without any direct risk of accident. 

Mahadevan \etal \cite{mahadevan2017communicating} investigate various modalities of \textit{communication} such as audio, visual, motion, etc. The authors note that in the absence of an explicit \textit{intent display} mechanism, pedestrians rely on \textit{vehicle speed and distance} to make crossing decision. As for different means of \textit{communication}, pedestrians generally prefer LED sequence signals to LCD displays and other modalities of communication such as auditory and physical cues. The authors show that the use of human-like features for communication such as animated faces on displays was not well-received by the participants. Overall, the authors recommend that  a combination of modalities including visual, physical and auditory should be considered. They point out that there is no limit on where the informative cues are located and can be either on the vehicle or in the environment. It should be noted that although this study is very thorough in terms of evaluating different design approaches, its scope is very limited. Only 10 subjects participated in the final phase of the study (Wizard of Oz phase) and the participants were all North American. Furthermore, the authors admit that \textit{culture} can play a very important role in the modality and type of communication preference.   

Implicit forms of communication such as vehicle's motion pattern  (\textit{speed and distance}) have also been investigated. Zimmerman \etal \cite{zimmermann2017first} show that abrupt acceleration behavior and short stopping distance by autonomous vehicles can be perceived as erratic behavior by pedestrians and negatively influence their crossing decision. The authors suggest that to be effective, a well-balanced acceleration and deceleration with sufficient distance to other road users should be used by autonomous vehicles. In another study Beggiato \etal \cite{beggiato2017right} examine the effect of vehicle's braking action whereby the vehicle can communicate its intention to pedestrians. The authors argue that the interpretation of the signal may vary with respect to other factors such as \textit{time of day}, \textit{vehicle speed}, and \textit{age}. For instance, older pedestrians generally make more conservative crossing decisions when the \textit{vehicle speed} is lower. 

Moving away from  \textit{communication},  Deb \etal \cite{deb2017development}, and similarly Hulse \etal \cite{hulse2018perceptions}, argue that the \textit{perceived risk} of autonomous vehicles may vary depending on pedestrians' \textit{age}, \textit{gender}, \textit{past experience}, level of \textit{law compliance}, \textit{location}, and \textit{social norms}. For example, younger male pedestrians, people with higher acceptance for innovation and people living in urban environments are more receptive of autonomous driving  technology. People with  traffic violation history also tend to be more comfortable when crossing in front of autonomous vehicles.   

\begin{figure}[!t]
\centering
\captionsetup[subfigure]{labelformat=empty}
\includegraphics[width=0.23\textwidth]{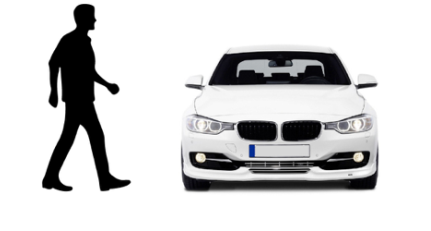}
\includegraphics[width=0.23\textwidth]{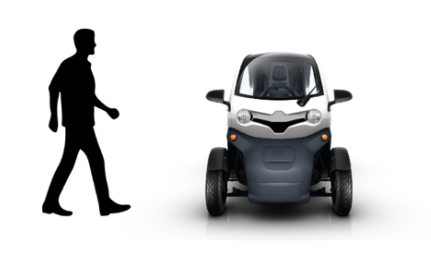}
\caption[dey Exp]{The vehicles used in \cite{dey2017impact}, an aggressive looking BMW (\textit{left}) and a friendly looking Renault (\textit{right}).}
\vspace*{-0.5cm} 
\label{fig:dey_exp}
\end{figure}

Dey \etal \cite{dey2017impact} evaluate the impact of \textit{vehicle type} on the \textit{perceived risk} of autonomous vehicles. The authors use two different types of vehicles, a BMW with an aggressive look and a Renault with a friendlier look (see \figref{fig:dey_exp}). They report that the \textit{vehicle speed and distance} compared to \textit{vehicle size} and \textit{appearance} play a more dominant role in crossing decision. Apart from dynamic factors, roughly 30\% of the participants claimed that they merely relied on the behavior of the car when making crossing decision, whereas the rest mentioned that \textit{vehicle size} was important to them rationalizing that the smaller the vehicle, the higher their chance of moving out of its way. The majority of the participants agreed that the friendliness of the vehicle design did not factor in their decision-making process.

Evaluating the impact of autonomous vehicle behavior on pedestrian crossing, Jayaraman \etal \cite{jayaraman2018trust}, argue that the presence of traffic \textit{signals} at crosswalks has little impact on pedestrian crossing decision and is highly determined by autonomous vehicle's driving behavior. The implication of these findings, however, is limited because the evaluation was performed only in a virtual reality environment.

\begin{figure*}[!t]
\centering
\includegraphics[width=0.8\textwidth]{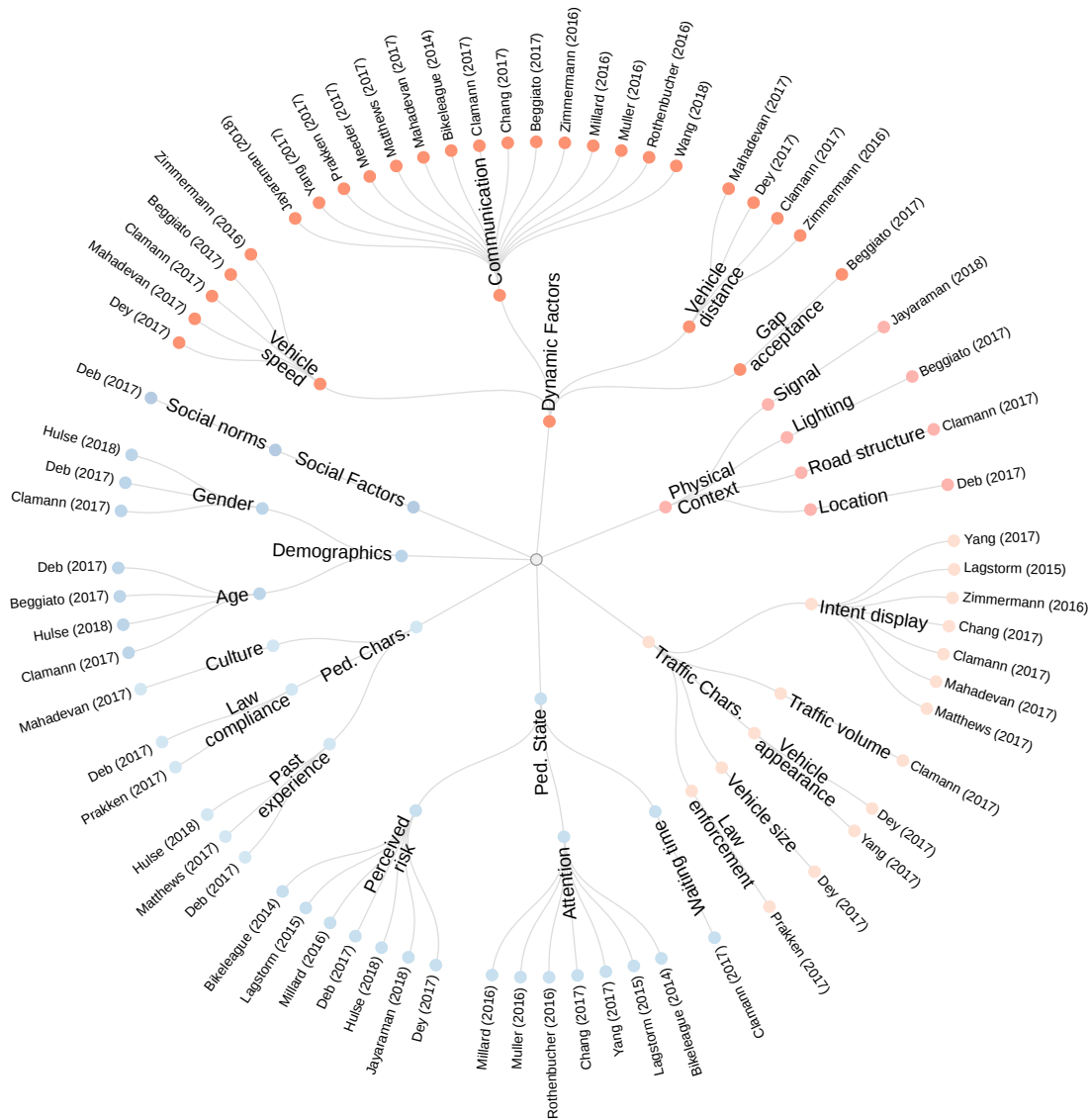}
\caption[factors and papers.]{A circular dendrogram of the factors influencing pedestrian behavior and the autonomous driving studies that identified them. Leaf nodes represent the individual studies (identified by the first author and year of publication) and internal nodes represent minor and major factors.}
\vspace*{-0.4cm} 
\label{fig:ped_factors_auto}
\end{figure*}

\figref{fig:ped_factors_auto} summarizes all of our findings on pedestrian behavior studies involving autonomous vehicles. At first glance, we can see that, compared to classical studies, pedestrian behavior in the context of autonomous driving is fairly understudied. The majority of research currently focuses on the role of \textit{communication}, \textit{intent display}, \textit{perceived risk} and \textit{attention}, while factors such as \textit{signal}, \textit{location}, \textit{road structure}, \textit{gap acceptance}, and \textit{social norms} are rarely addressed. More importantly, some of the factors widely studied in classical works, namely \textit{group size}, \textit{pedestrian speed}, and \textit{street width}, have not been evaluated in the context of autonomous driving.  

\section{Interaction Between Pedestrians and Autonomous Vehicles: Practical Approaches} 
\subsection{Communicating with Pedestrians}

As mentioned in the previous section, the changes in motion can be used as one of the means of communication between pedestrian and autonomous vehicles \cite{zimmermann2017first,risto2017human}. Here, however, we focus on explicit forms of communication some of which were discussed earlier.

One way of direct communication with traffic is via radio signals. Vehicle to Vehicle (V2V) and Vehicle to Infrastructure (V2I), which are collectively known as V2X (or Car2X in Europe), are examples of such technologies \cite{hobert2015enhancements,cheng2014index}. These methods are essentially a real-time short-range wireless data exchange between the entities allowing them to share information regarding their pose, speed, and location \cite{narla2013evolution}.

Recent developments extend the idea of V2X communication to connect Vehicles to Pedestrians (V2P). For instance, Honda proposes to use pedestrians' smartphones to broadcast their whereabouts as well as to receive information regarding the vehicles in their proximity. Using this method, both smart vehicles and pedestrians are aware of each other's movements, and if necessary, receive warning signals when an accident is imminent \cite{cunningham2013}. Hussein \etal \cite{hussein2016p2v} propose the use of a smartphone application that broadcasts the position of the pedestrian and receives the location of nearby autonomous vehicles. The application then calculates and predicts the location and time of the collision, and if the pedestrian is in danger, sends a warning signal. Gordon \etal \cite{gordon2016automated} patented a wearable sensor technology for pedestrians to receive warning signals from autonomous vehicles. 

In spite of their effectiveness in preventing accidents,  V2P technologies raise a number of concerns one of which is the privacy issues associated with sharing road users' personal information \cite{schmidt2015v2x}. Moreover, studies show that a large number of pedestrians are reluctant to use V2P technologies claiming that these shift the responsibility of potential accidents to pedestrians and away from autonomous vehicles \cite{bikeleauge2014}.

\begin{figure}[!t]
\centering

\subfloat[]{\includegraphics[width=0.22\textwidth, height=2cm]{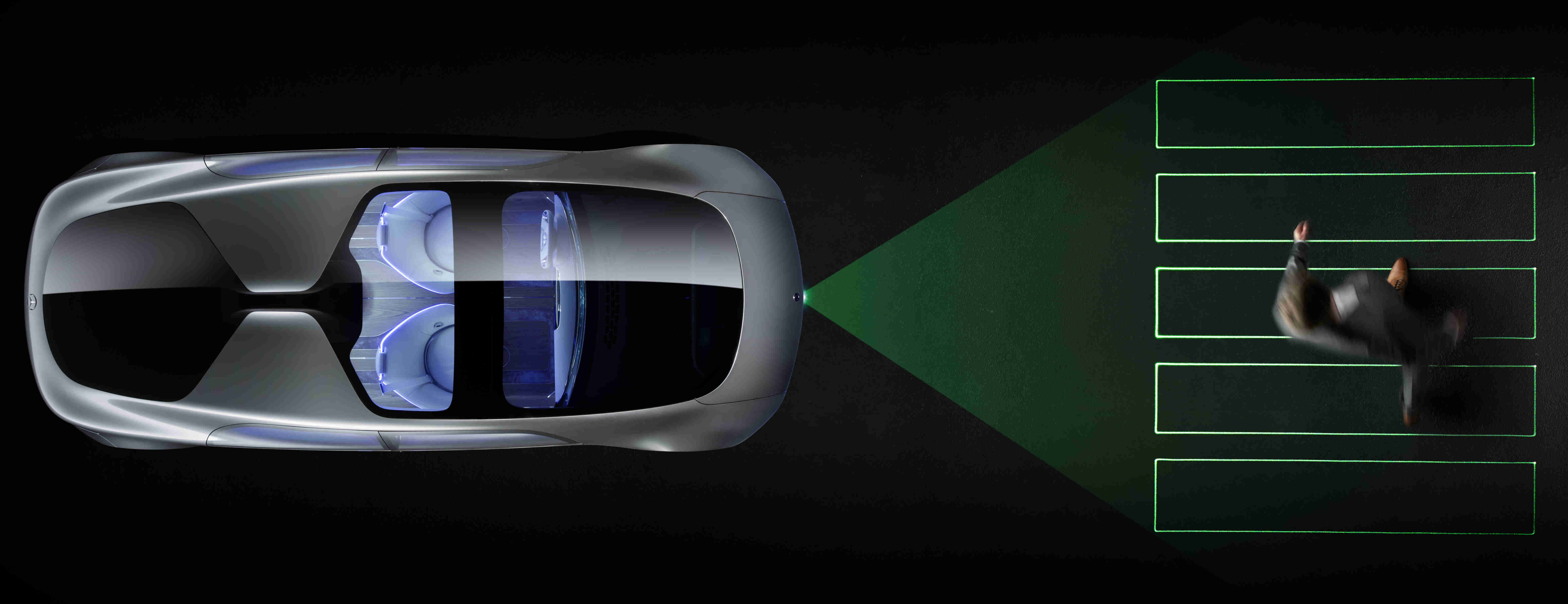}
\label{fig:benz_auto}}
\hspace{0.2cm}
\subfloat[]{\includegraphics[width=0.22\textwidth, height=2cm]{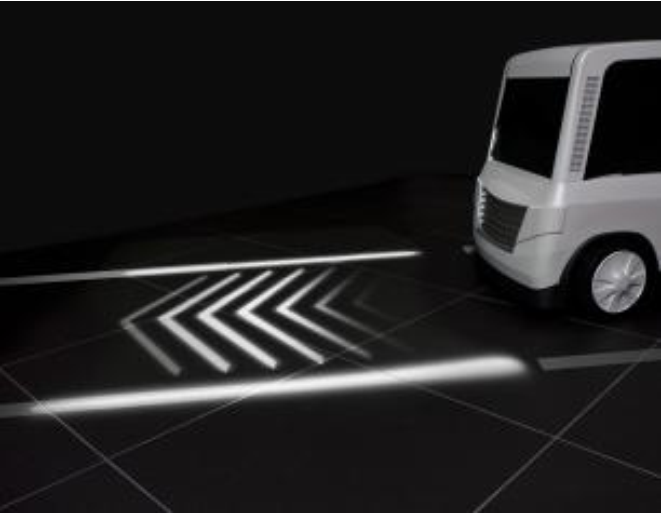}
\label{fig:mitsubishi}}\\
\subfloat[]{\includegraphics[width=0.22\textwidth, height=2cm]{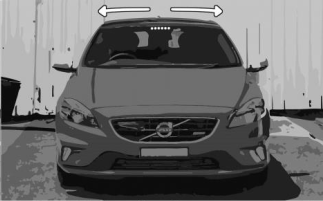}
\label{fig:avip}}
\hspace{0.2cm}
\subfloat[]{\includegraphics[width=0.22\textwidth, height=2cm]{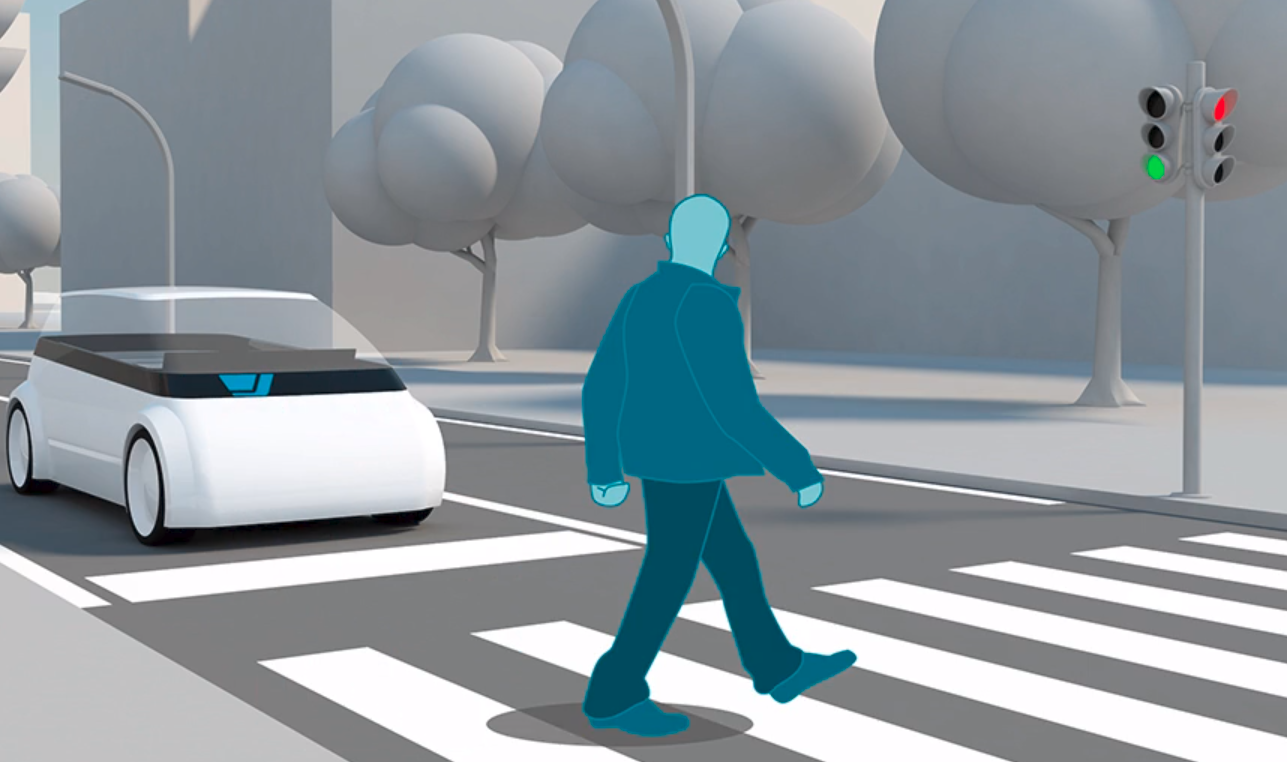}
\label{fig:autonomi}}\\
\subfloat[]{\includegraphics[width=0.22\textwidth, height=2cm]{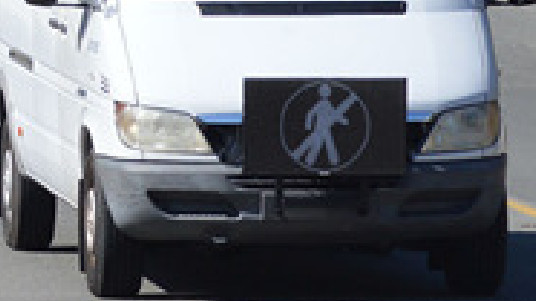}
\label{fig:display}}
\hspace{0.2cm}
\subfloat[]{\includegraphics[width=0.22\textwidth, height=2cm]{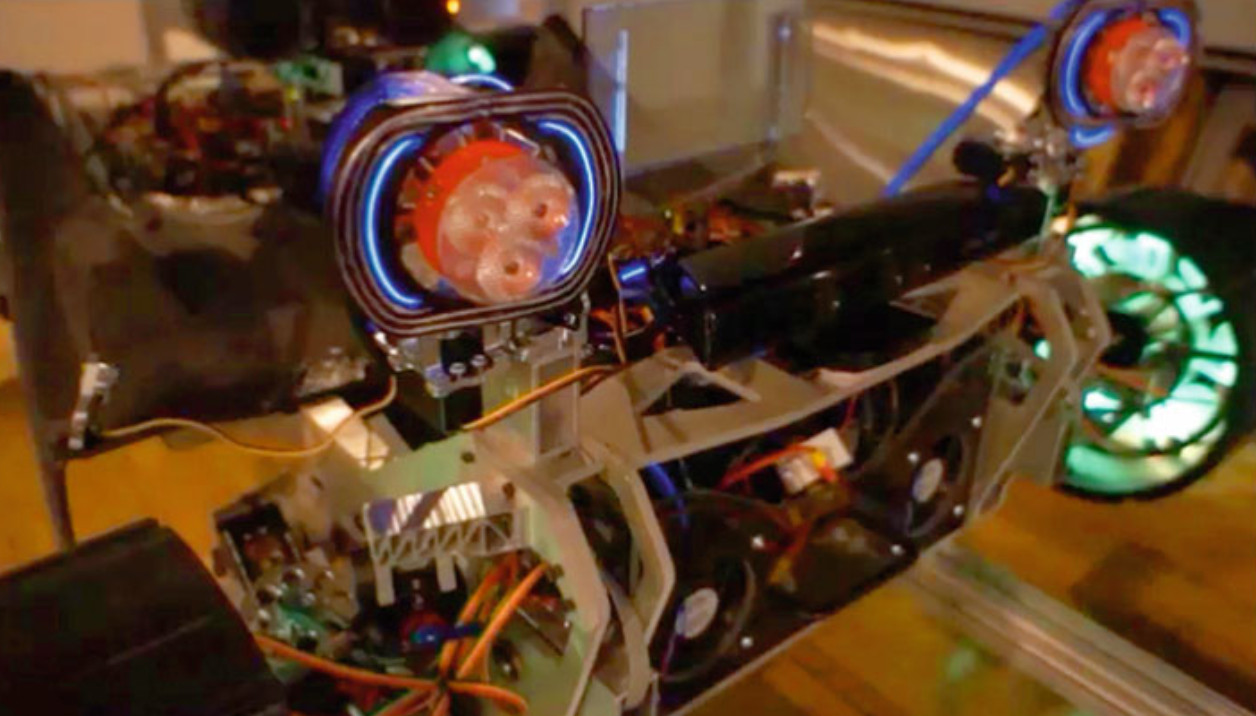}
\label{fig:aevita}}\\
\caption[Comm methods]{Different concepts of communication for autonomous vehicles. a) Mercedes-Benz zebra crossing projection \cite{benzauto}, b) Mitsubishi forward indicator \cite{mitslights}, c) LEDs indicating yield \cite{lagstrom2015avip}, d) AutonoMI pedestrian detection and tracking indicator \cite{graziano2014}, e) advisory display \cite{clamann2017evaluation}, and f) AEVITA moving eye concept \cite{pennycooke2012aevita} (source \cite{farber2016communication}).}
\vspace*{-0.5cm} 
\label{fig:cars_communicate}
\end{figure}

Recent research has been focusing on different modalities of communication. The use of displays is a common technique to transmit a message  \cite{googledisp,zimmermann2017first,mahadevan2017communicating}. Such displays can either transmit informative messages, for instance, the speed of the vehicle \cite{clamann2017evaluation} or intention of the vehicle \cite{matthews2017intent}, or they can be advisory, meaning that they suggest a course of action to pedestrians, e.g. a sign indicating cross/not to cross \cite{clamann2017evaluation,smartcomm} (see \figref{fig:display}). 

In \cite{lagstrom2015avip} the authors recommend the use of an array of LED lights on top of the windshield (\figref{fig:avip}) to transmit messages. For example, when the middle lights are on, it means the vehicle is in autonomous mode, and various lighting up patterns indicate whether the car is yielding or is about to move. An LED-like display, called AutonoMI, is proposed by Graziano \cite{graziano2014}. When the vehicle encounters a pedestrian, the part of the LED array closest to the pedestrian lights up, acknowledging that the pedestrian is recognized. When the pedestrian begins crossing, the array follows the pedestrian to assure them that they are still being seen (see \figref{fig:autonomi}).

A combination of LEDs with other communication modalities have been investigated. Florentine \etal \cite{florentine2016pedestrian} use color LEDs in conjunction with an audio module to cast warning signals. Siripanich \cite{siripanich2017} combines LED lights with advisory signs to simultaneously inform and advise pedestrians. In addition to LED lights and audio signals, Mahadevan \etal \cite{mahadevan2017communicating} recommend the use of a physical signal such as a moving robotic hand attached to the vehicle.  

Informative signals regarding the intention of the vehicle can also be displayed on the road surface using projectors \cite{hillis2016communication,mitslights}. Mitsubishi \cite{mitslights} introduces road-illuminating directional indicator which projects large, easy-to-understand animated illuminations on road surfaces indicating the intention of the vehicle such as forward or reverse driving (see \figref{fig:mitsubishi}). Mercedes-Benz, in their most recent concept autonomous vehicle (as illustrated in \figref{fig:benz_auto}), uses a combination of techniques including series of LED lights at the rear end of the car to ask other vehicles to stop/slow or inform them if a pedestrian is crossing, a set of LED fields at the front to indicate whether the vehicle is in autonomous or manual mode and a projector that can project zebra crossing on the ground for pedestrians \cite{benzauto}.

To make the communication with pedestrians more human-like, some researchers propose the use of human-like eyes on vehicles \cite{chang2017eyes,pennycooke2012aevita}. For example, a moving-eyes approach is used in \cite{pennycooke2012aevita} in which the vehicle is able to detect the gaze of the pedestrians, and, using rotatable front lights, it establishes (the feeling of) eye contact with the pedestrians and follow their gaze (see \figref{fig:aevita}). Some researchers also go so far as suggesting to use a humanoid robot in the driver seat to perform human-like gestures or body movements during communication \cite{mirnig2017three}.

\begin{figure*}[!t]
\centering
\captionsetup[subfigure]{labelformat=empty}

\includegraphics[width=5.75cm,height=3.25cm]{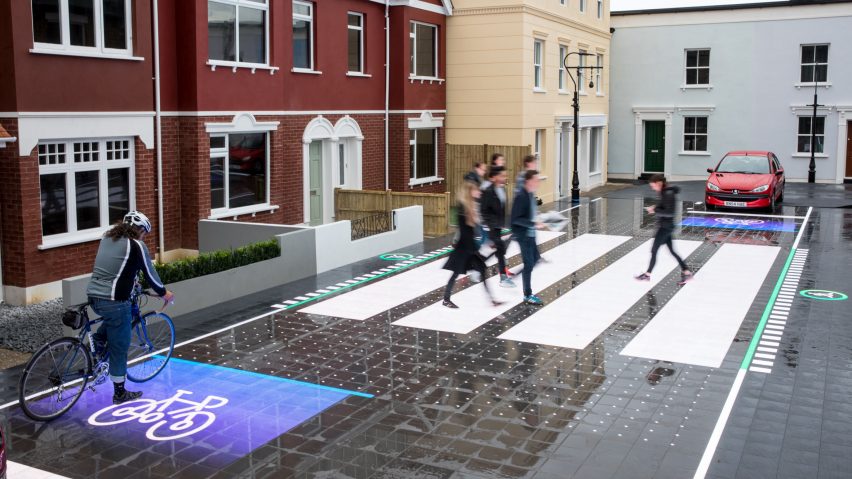}
\label{fig:smart_cross}
\includegraphics[width=5.75cm,height=3.25cm]{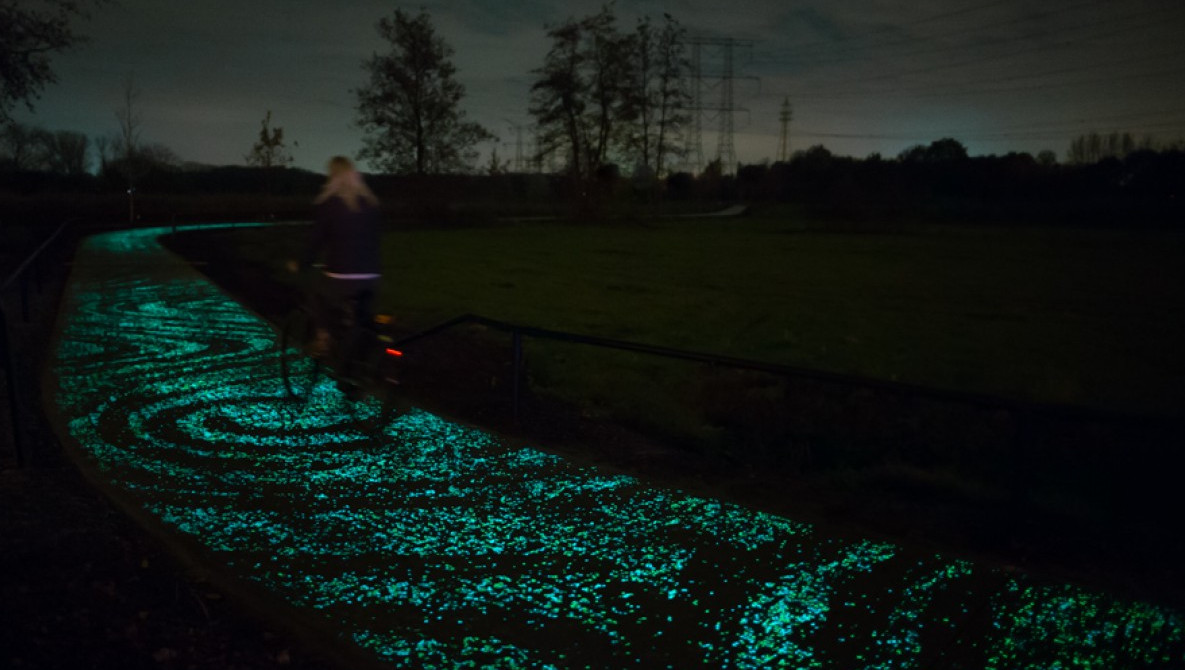}
\label{fig:smart_vangogh}
\includegraphics[width=5.75cm,height=3.25cm]{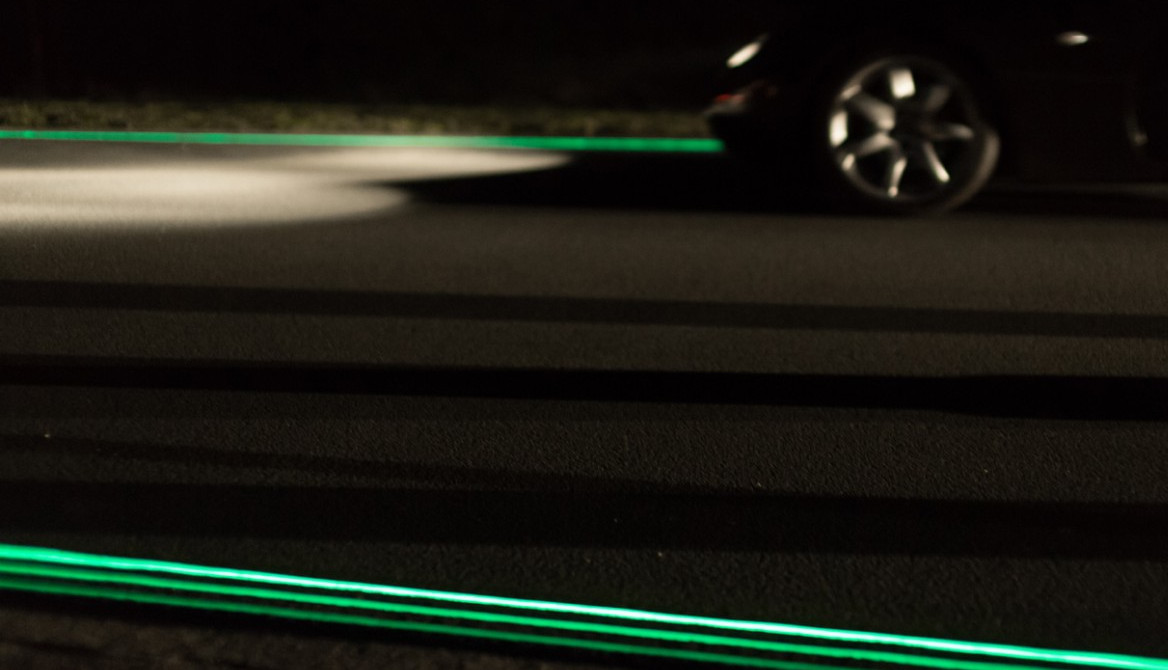}
\caption[smart infrastructure]{Examples of smart road concept. \textit{from left to right} Umbrellium smart crossing \cite{mairs2017road}, and Studio Rosegaarde Van Gogh path and highway glowing lines \cite{smartroad}.}
\vspace*{-0.4cm} 
\label{fig:smart_roads}
\end{figure*}

Roadways can also be used to transmit the intentions and whereabouts of the road users. During recent years, the concept of smart roads has been gaining popularity in the field of intelligent driving. Smart roads are equipped with sensors and lighting equipment, which can sense events such as vehicle or pedestrian crossing, changes in weather conditions or various hazards that can potentially result in accidents. Through the use of visual effects, the roads then inform the road users about the potential threats \cite{siess2015hybrid}.

Today, a few instances of smart roads have been implemented. Last year Umbrellium unveiled a new interactive crossing in London equipped with LEDs which flash various warning signals to distracted road users or display zebra crossing lines for pedestrians \cite{mairs2017road}. Studio Rosegaarde \cite{smartroad} implemented various types of smart roads in Netherlands such as the Van Gogh path which highlights traversable paths for pedestrians or glowing lines which highlights the boundaries of highways at night (see \figref{fig:smart_roads}).

\subsection{Understanding Pedestrians' Intentions}

In intelligent driving systems, intention estimation techniques have been widely used for predicting the behavior of the drivers \cite{ohn2014head,molchanov2015multi}, other drivers \cite{laugier2011probabilistic,li2016recognizing}, pedestrians \cite{kooij2014analysis,kohler2012early} or combinations of any of these three \cite{phan2013estimating,bahram2016combined} (for a more detailed list of these techniques see \cite{ohn2016looking}). In this section, however, we only discuss the pedestrian intention estimation methods in the context of intelligent transportation systems mentioning a few techniques used in mobile robotics.

Typically, intention estimation algorithms are very similar to object tracking systems. One's intention can be estimated by looking at their past and current behavior including their dynamics, current activity and context. 

There are a number of works that purely rely on data meaning that they attempt to model pedestrian walking direction with the assumption that all relevant information is known to the system. These models either base their estimation on dynamic information such as the position and velocity of pedestrians \cite{schulz2015controlled}, or in addition, take into account the contextual information of the scene such as pedestrian signal state, whether the pedestrian is walking alone or in a group, and their distance to the curb \cite{hashimoto2015probability}. In a work by Brouwer \etal \cite{brouwer2016comparison}, the authors investigate the role of different types of information in collision estimation. More specifically, they consider the following four factors: \textit{dynamics} (directions pedestrian can potentially move to and time to collision), \textit{physical elements} (pedestrian's moving direction and distance to the car and velocity), \textit{awareness} (in terms of head orientation towards the vehicle), and \textit{obstacles}. The authors show that, in isolation, \textit{physical elements} and \textit{awareness} are the best predictors of collisions, and combining all four factors together, the best prediction results can be achieved.

Vision-based intention estimation algorithms often treat the problem as tracking a dynamic object by taking into account the changes in the position, velocity and orientation of pedestrians \cite{goldhammer2014analysis,trivedi2015trajectory} or by considering the changes in their 3D pose \cite{quintero2014pedestrian}. For instance, in \cite{vcermak2017learning}, the authors use a neural network architecture to make a binary \enquote*{stop/go} decision given the current position of pedestrians. Kooij \etal \cite{kooij2014analysis} employ a dynamical Bayesian model, which takes as input the current position of the pedestrian and, based on their motion history, infers in which direction the pedestrian might move next. In addition to pedestrian position, Volz \etal \cite{volz2016data} use information regarding the pedestrian's distance to the curb and the car as well as the pedestrian's velocity at the time. This information is fed into an LSTM network to infer whether the pedestrian is going to cross the street.

In robotics, intention prediction algorithms are used as a means of improving trajectory selection and navigation. Besides dynamic information, these techniques assume a potential goal for pedestrians based on which their trajectories are predicted \cite{bandyopadhyay2013intention,bai2015intention}.

\begin{table*}[!hbtp]
\caption[A summary of intention estimation algorithms.]{A summary of intention estimation algorithms. Abbreviations: \textit{Factors:} PP = Pedestrian Position, PV = Pedestrian Velocity, SC = Social Context, PPs = Pedestrian Pose, SS = Street Structure, MH = Motion History, HO = Head Orientation, G = Goal, GS = Group Size, Si = Signal, DC = Distance to curb, DCr = Distance to Crosswalk, DV = Distance to Vehicle, VD = Vehicle Dynamics, \textit{Inference:} GD = Gradient Descent, PF = Particle Filter, GP = Gaussian Process, NN = Neural Network, \textit{Prediction Type:} Traj = Trajectory, Cross = Crossing, \textit{Data Type:} Img = Image, Col = Color, Vid = Video, Gr = Grey, St = Stereo, I = Infrared, \textit{Camera Position:} F = Front view, BeV = Bird's Eye View, Mult = Multiple views, FP = Fixed Position on-site.}
\centering
\resizebox{0.8\textwidth}{!}{
\begin{tabular}{|c|c|c|c|c|c|c|}
\hline
Model&Year&Factors&Inference&Pred. Type&Data Type&Cam Position\\
\hline
\hline
LTA \cite{pellegrini2009you}&2009&PP,PV,G,SC&GD&Traj&Vid+Col&BeV+FP\\ \hline
Early-Det \cite{kohler2012early}&2012&PPs&SVM&Cross&Img+Col&F+FP\\ \hline
IAPA \cite{bandyopadhyay2013intention}&2013&PP,PV,G&MDP&Traj,Cross&Vid+Col&F\\ \hline
Evasive \cite{kohler2013autonomous}&2013&PPs,MH&SVM&Cross&St+Vid+Gr&F+FP\\ \hline
Early-Pred \cite{goldhammer2013early}&2014&PP,PV&SVM&Traj&Vid+Gr&Mult+FP\\ \hline
Veh-Perspective \cite{kooij2014analysis}&2014&PP,PV,VD&BN&Traj&Vid+Gr&F\\ \hline
Context-Based \cite{kooij2014context}&2014&PP,SS,HO,VD&BN&Cross&Vid+Gr &F\\ \hline
Intent-Aware \cite{madrigal2014intention}&2014&PP,PV,SC&BN&Traj&Vid+Col&F\\ \hline
Path-Predict \cite{quintero2014pedestrian}&2014&PP,PV,PPs&BN&Traj,Pose&Vid+Col&F\\ \hline
MMF \cite{schulz2015controlled}&2015&PP,PV,SC&CRF&Traj&Vid+Gr&F\\ \hline
Intend-MDP \cite{bai2015intention}&2015&PP,PV,G,VD&MDP&Traj&Vid+Col+L&F\\ \hline
SVB \cite{kohler2015stereo}&2015&PPs,MH&SVM&Cross&St+Vid+Gr&F+FP\\ \hline
PE-PC \cite{hashimoto2015probability}&2015&PP,PV,GS,SI&BN&Traj,Cross&Vid+Gr&F\\ \hline
Traj-Pred \cite{trivedi2015trajectory}&2015&PP,PV&PF&Traj&Vid+Col&F\\ \hline
FRE \cite{volz2015feature}&2015&PV,DC,DCr,DV,VD&SVM&Cross&Vid+L&F\\ \hline
Eval-PMM \cite{brouwer2016comparison}&2016&PP,PV,HO&BN&Traj&Vid+Col&F\\ \hline
ECR \cite{hariyono2016estimation}&2016&PP,PV,HO &BN&Collision&Vid+Col+Gr&F\\ \hline
HI-Robot \cite{park2016hi}&2016&MH&GP&Collision&Vid+L&F\\ \hline
CBD \cite{schneemann2016context}&2016& PP,SS,MH&SVM&Cross&Vid+Col&F\\ \hline
DDA \cite{volz2016data}&2016&DC,DCr,DV,VD&NN&Cross&Vid+L&F\\ \hline
DFA \cite{kwak2017pedestrian}&2017&PP,PV,MH&DFA&Cross&Vid+I&F\\ \hline
Cross-Intent \cite{rasouli2017they}&2017&PPs,SS,HO&NN,SVM&Cross&Vid+Col&F\\ \hline
Proxy-Learn \cite{vcermak2017learning}& 2017& PP& NN& Collision& Img+Col& F\\ \hline
Ped-Phones \cite{rangesh2018vehicles}&2018&PPs&SVM,BN&Pose&Vid+Col&F\\ \hline
\end{tabular}
}
\vspace*{-0.4cm} 
\label{table:intention_est}
\end{table*}

Merely relying on pedestrian trajectory and dynamic factors in estimation one's intention is subject to error. For example, pedestrians may start walking suddenly, change their direction abruptly or stop. Moreover, observed pedestrians may be stationary or even walk alongside the street while checking on traffic to cross. In such scenarios, a trajectory-based algorithm may flag the pedestrians as no collision threat even though they might be crossing shortly \cite{schmidt2009pedestrians}.

\begin{figure}[!tp]
\centering
\includegraphics[width=0.9\columnwidth]{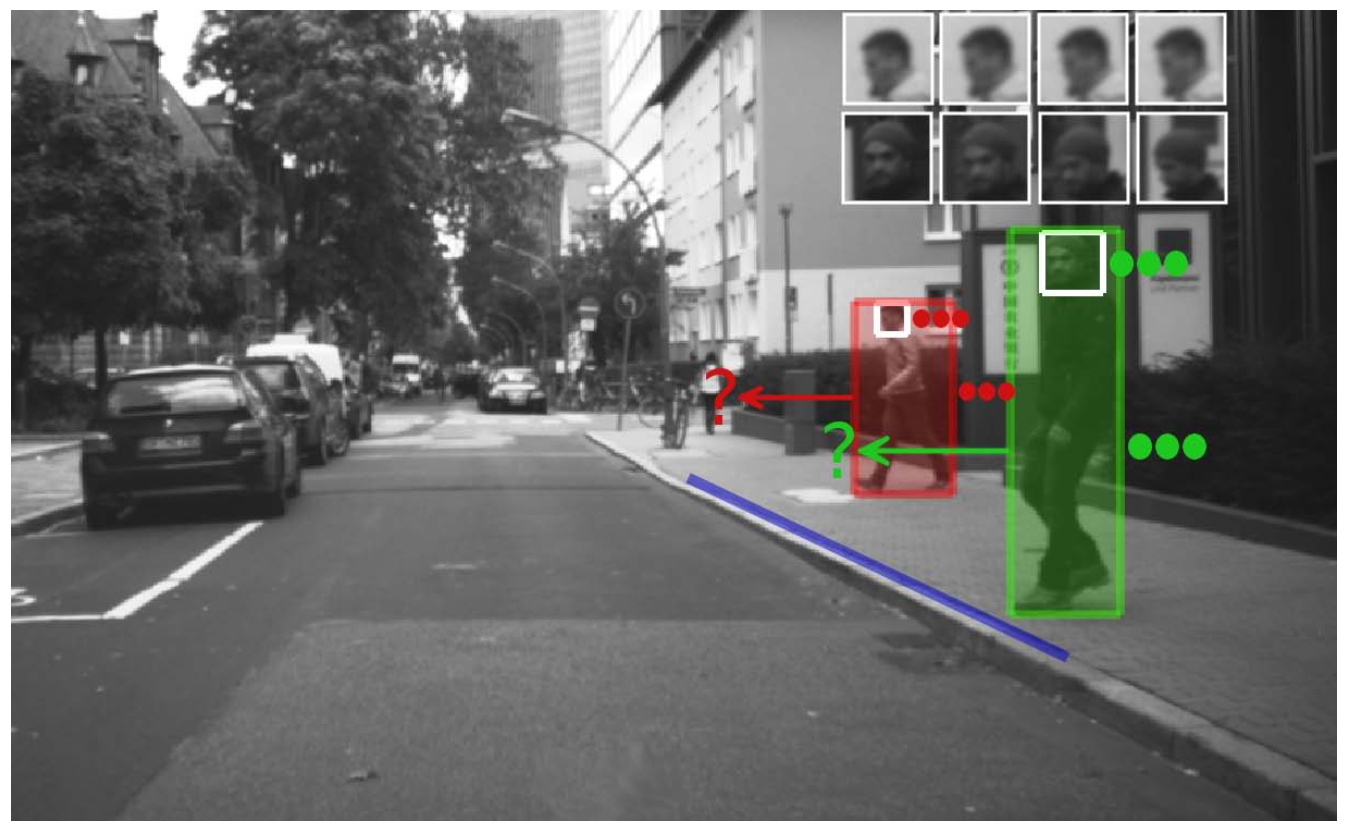}
\caption[smart]{An example of pedestrian intention estimation using contextual cues. Source: \cite{kooij2014context}.}
\vspace*{-0.5cm} 
\label{fig:intention_est}
\end{figure}

In some recent works, social context is exploited to estimate intention and deal with shortcomings of trajectory-based approaches. For instance, pedestrian awareness is measured by pedestrians' head orientation relative to the vehicle \cite{kooij2014context,hariyono2016estimation,kwak2017pedestrian}. Kooij \etal \cite{kooij2014context} employ a graphical model that takes into account factors such as pedestrian trajectory, distance to the curb and awareness (see \figref{fig:intention_est}). Here, they argue that the pedestrian looking towards the car is a sign that they noticed the car and is less likely to cross the street. This model, however, is based on data collected from a scripted experiment which means that the participants were instructed to perform certain actions, and all videos were recorded in a narrow non-signalized street.

For intention estimation, social forces, which refer to people's tendency to maintain a certain distance from one another, are also considered. In their simplest form, social forces can be treated as a dynamic navigation problem in which pedestrians choose the path that minimizes the likelihood of colliding with others \cite{pellegrini2009you}. Social forces also reflect the relationship between pedestrians, which in turn can be used to predict their future behavior. For instance, Madrigal \etal \cite{madrigal2014intention} define two types of social forces: repulsion and attraction. In this interpretation, for example, if two pedestrians are walking close to one another for a period of time, it is more likely that they are interacting, therefore the tracker estimates their future states close together.

Apart from the explicit tracking of pedestrian behavior, a number of works try to solve the intention estimation problem using various classification approaches. Kohler \etal \cite{kohler2012early}, via an SVM algorithm, classify pedestrian posture as \enquote*{about to cross} or \enquote*{not crossing}. The postures are extracted in the form of silhouette body models from motion images generated by background subtraction. In the extensions of this work \cite{kohler2013autonomous,kohler2015stereo}, the authors use a HOG-based detection algorithm to first localize the pedestrian, and then, using stereo information, to extract the body silhouette from the scene. To account for the previous action, they perform the same process for N consecutive frames and superimpose all silhouettes into a single image. The final image is used to classify whether the pedestrian is going to cross.

Rangesh \etal \cite{rangesh2018vehicles} estimate the pose of pedestrians in the scene, and identify whether they are holding cell phones. The combination of the pedestrians' pose and the presence of a cellphone is used to estimate the level of pedestrians engagement in their devices. In \cite{rasouli2017they}, the authors use various contextual information such as characteristics of the road, the presence of traffic signals and zebra crossing lines in conjunction with pedestrians' state to estimate whether they are going to cross. In this method, two neural network architectures are used. One network is responsible for detecting contextual elements in the scene and the other for identifying whether the pedestrian is walking/standing and looking/not-looking. The scores from both networks are then fed to a linear SVM to classify the intention of the pedestrians. The authors report that by taking into account the context, intention estimation accuracy can be improved by up to 23\%.   

Schneemann \etal \cite{schneemann2016context} consider the structure of the street as a factor influencing crossing behavior. The authors generate an image descriptor in the form of a grid which contains the following information: \textit{street-zones} in the scene including ego-zone (the vehicle's lane), non-ego lanes (other street lanes), sidewalks, and mixed-zones (places where cars may park), \textit{crosswalk occupancy} (the position of scene elements with respect to the current position of the pedestrians), and \textit{waiting area occupancy} (occupancy of waiting areas such as bus stops with respect to the pedestrian's orientation and position). Such descriptors are generated for a number of consecutive frames and concatenated to form the final descriptor. At the end, an SVM model is used to decide how likely the pedestrian is to cross. Despite its sophistication for exploiting various contextual elements, this algorithm does not perform any perceptual tasks to identify the aforementioned elements and simply assumes they are all known in advance.

In the context of robotic navigation, Park \etal \cite{park2016hi} classify observed pedestrian trajectories to measure the imminence of collisions. The authors recorded over 2.5 hours of videos of the pedestrians who were instructed to engage in various activities with the robot (e.g. approaching the robot for interaction or simply blocking its way). Using a Gaussian process, the trajectories were then classified into blocking and non-blocking.

Table \ref{table:intention_est} gives a summary of the papers discussed in this section. Overall, there is no particular trend in the type of information (e.g. pedestrian dynamics or contextual information) utilized for estimating pedestrian crossing decision. One possible reason could be the availability and type of data used for training intention estimation algorithms.

To date, there are very few publicly available datasets that are tailored to pedestrian intention estimation applications. Pedestrian detection datasets such as Caltech \cite{dollarCVPR09peds} or KITTI \cite{Geiger2013IJRR} are often used for predicting crossing behavior. These datasets contain a large number of pedestrian samples with bounding box annotations and temporal correspondences allowing one to detect and track pedestrians in multiple frames. Some datasets also have added contextual information particularly for pedestrian crossing behavior understanding. For instance, Daimler-Path \cite{schneider2013pedestrian} and Daimler-Intent \cite{kooij2014context} contain pedestrian head orientation information. A more recent dataset, JAAD \cite{rasouliagree}, in addition to a large number of pedestrian samples (over 2700) with bounding boxes, is annotated with detailed contextual information, e.g. weather condition, street structure, and delineation, as well as pedestrian characteristics and behavioral information, e.g. demographics, group size, pedestrian state and communication cues.

\vspace{-0.3cm}

\section{what's next}

In this section, we will discuss open problems mentioned in the paper thus far. 

\subsubsection{Classical studies of pedestrian behavior} We identified 38 factors that can potentially  impact the way pedestrians behave. Some of these factors have been studied more than the others (see \figref{fig:ped_factors_all}) such as age, gender, group size and gap acceptance. In the literature, there is a consensus about the influence of the majority of these factors, for example, how group size influences gap acceptance or how individuals behave based on their demographics.

However, often the results presented by these studies are contradictory especially the ones on topics such as communication, the influence of imitation, the role of attention, waiting time influence on gap acceptance, etc. Although some of these contradictions can be explained by the differences in the methods of studies, we believe that  the main reason is the variations in \textit{culture}, \textit{time of study} and \textit{interrelationships} between the factors. 

Culture can influence pedestrian behavior in many ways. The studies often are conducted in different geographical locations where culture and social norms can be quite different. This means a number of these studies have to be reproduced in different regions to account for cultural differences.   

Changes in socioeconomic and technological factors also influence traffic behavior. For example, compared to the 1950s or 1960s, today vehicles are much safer, roads are built and maintained better, the number of vehicles and pedestrians have increased significantly, and traffic laws have been changed, all of which change traffic dynamics. To account for modern time pedestrian behavior, some of these studies have to be repeated.

As illustrated in \figref{fig:ped_factors_relationship}, there is a strong interrelationship between factors that influence pedestrian behavior. This means that only studying a small subset of these factors may not capture the true underlying reasons behind pedestrian crossing decision. Therefore to avoid fallacies when reasoning about pedestrian behavior, studies have to be multi-modal and account for chain effects that factors might have on each other.

\subsubsection{Pedestrian behavior and autonomous vehicles} 

In recent years behavioral studies in the context of autonomous vehicles have gained momentum resulting in a number of published works on pedestrian behavior towards autonomous cars. The number of these studies, however, is still relatively small, compared to classical studies. Although classical studies have a number of implications for autonomous driving systems, it is reasonable to expect that pedestrians might behave differently when facing autonomous vehicles. This means more studies of similar nature to classical studies have to be conducted involving autonomous vehicles. 

The scope of the majority of behavioral studies involving autonomous vehicles is also very limited, both in terms of sample size (often less than 100) and demographics of participants (e.g. university students). As a result, some of these studies have reported very contradictory findings. To be useful for the design of autonomous vehicles, these works have to be conducted on a much larger scale, and of course, follow the same considerations as classical behavior studies.    

\subsubsection{Communicating with road users} 

Designing interfaces for autonomous vehicles in order to communicate with pedestrians is an ongoing research problem. One of the main questions to answer is what modality of communication is most effective. Unfortunately, the majority of the research in this field fails to address this issue. For example, some studies focus on whether any form of communication is important or compare different strategies within the same modality (e.g. informative vs advisory LCDs or how to light up LED lights). There are very few studies addressing communication mechanisms across different modalities, and if so, their empirical evaluation is limited to a sample size of no more than 10 participants. This points to the need for studies in a larger scale using human participants with diverse background.

\subsubsection{Understanding pedestrians' intention}

The current intention estimation algorithms are very limited in terms of using various contextual information and often do not involve necessary visual perception algorithms to analyze the scenes. In addition, data used in these algorithms is either scripted or not sufficiently diverse to include various traffic scenarios. To be effective, these algorithms should be able to, first, identify the relevant elements in the scene, second, reason about the interconnections between these elements, and third, infer the upcoming actions of the road users.

In addition, these systems should be universal in a sense that they can be used in various traffic scenarios with different street structures, traffic signals, crosswalk configurations, etc. To facilitate this objective, the first step is to collect behavioral data under various traffic conditions and from different geographical locations.

\vspace{-0.25cm}

\section*{Acknowledgment}
This work was supported by the Natural Sciences and Engineering
Research Council of Canada (NSERC), the NSERC Canadian
Field Robotics Network (NCFRN), the Air Force Office for Scientific Research (USA), and the Canada Research Chairs Program through grants to JKT.

\vspace{-0.20cm}

\Urlmuskip=0mu plus 1mu\relax
\bibliographystyle{IEEEtran}
\bibliography{tits_refs}

\begin{IEEEbiography}
[{\includegraphics[width=1in,height=1.25in,clip,keepaspectratio]{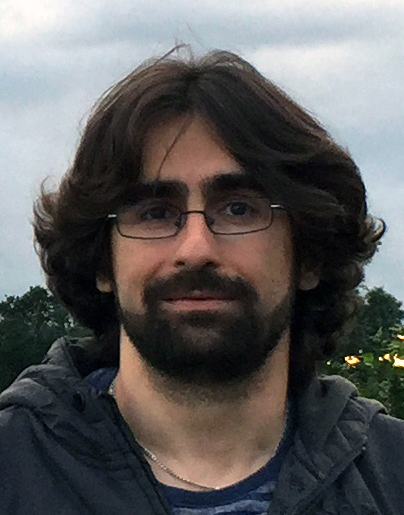}}]
{Amir Rasouli}received his B.Eng. degree in Computer Systems Engineering at Royal Melbourne Institute of Technology in 2010 and his M.A.Sc. degree in Computer Engineering at York University in 2015. He is currently working towards the PhD degree in Computer Science at the Laboratory for Active and Attentive Vision, York University. His research interests are autonomous robotics, computer vision, visual attention, autonomous driving and related applications.
\end{IEEEbiography}

\begin{IEEEbiography}
[{\includegraphics[width=1in,height=1.25in,clip,keepaspectratio]{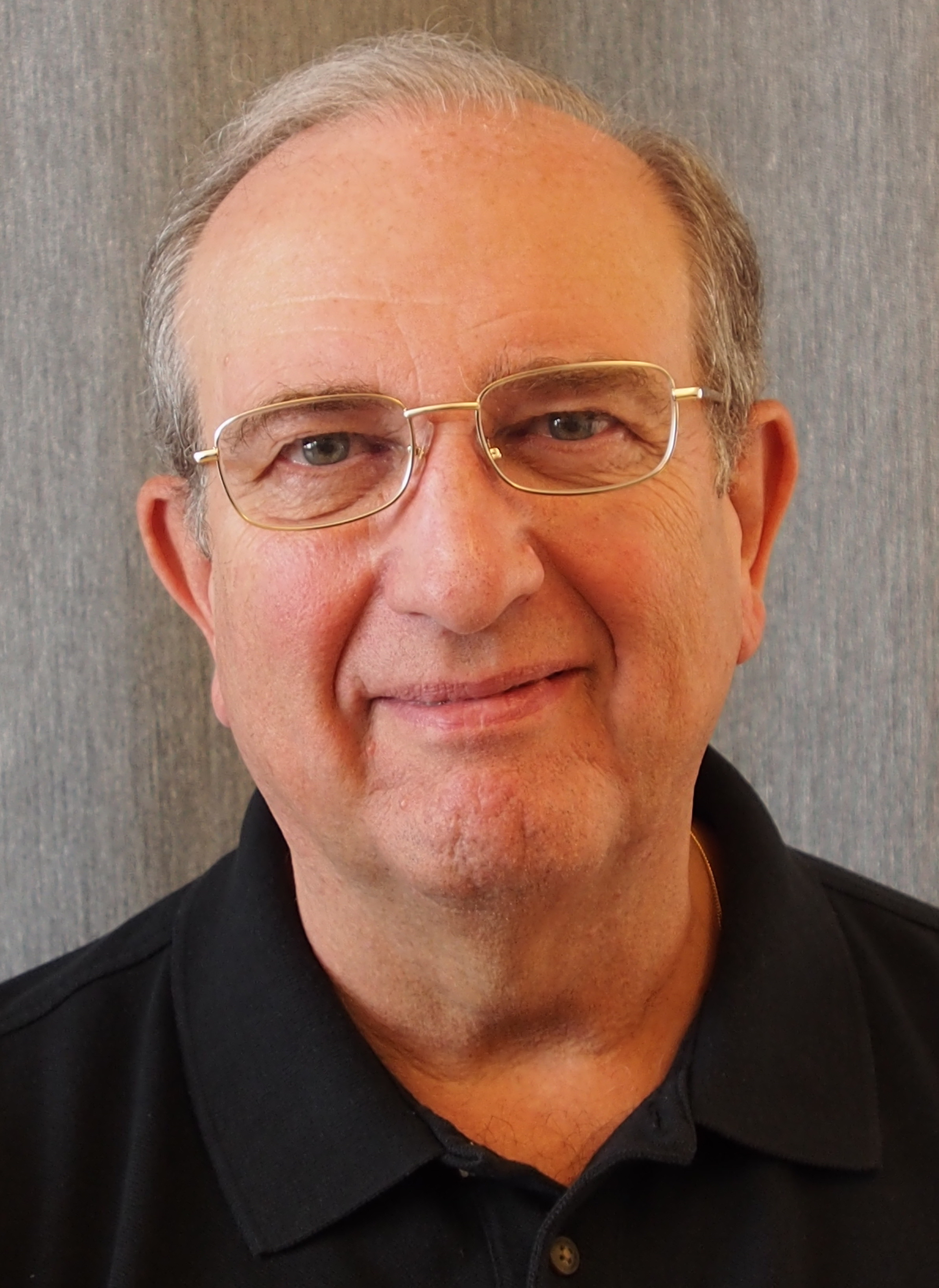}}]
{John K. Tsotsos} is Distinguished Research Professor of Vision Science at York University. He received his doctorate in Computer Science from the University of Toronto. After a postdoctoral fellowship in Cardiology at Toronto General Hospital, he joined the University of Toronto on faculty in Computer Science and in Medicine. In 1980 he founded the Computer Vision Group at the University of Toronto, which he led for 20 years. He was recruited to York University in 2000 as Director of the Centre for Vision Research. His current research focuses on a comprehensive theory of visual attention in humans. A practical outlet for this theory embodies elements of the theory into the vision systems of mobile robots. 
\end{IEEEbiography}

\end{document}